\documentclass[lettersize,journal]{IEEEtran}
\usepackage{amsmath,amsfonts}
\usepackage{array}
\usepackage[caption=false,font=normalsize,labelfont=sf,textfont=sf]{subfig}
\usepackage{textcomp}
\usepackage{stfloats}
\usepackage{url}
\usepackage{verbatim}
\usepackage{graphicx}
\usepackage{cite}
\def\BibTeX{{\rm B\kern-.05em{\sc i\kern-.025em b}\kern-.08em
    T\kern-.1667em\lower.7ex\hbox{E}\kern-.125emX}}
\usepackage{balance}

\usepackage{times}
\usepackage{epsfig}
\usepackage{amssymb}
\usepackage{booktabs}
\usepackage{algorithm}
\usepackage{algorithmicx}
\usepackage{tikz}
\usepackage{comment}
\usepackage{bbding}
\usepackage{mathrsfs}
\usepackage{algpseudocode}
\usepackage{pythonhighlight}
\usepackage{newfloat}
\usepackage{listings}
\usepackage{multirow}
\usepackage{tabularx}
\usepackage{makecell}
\usepackage{xcolor}

\begin{document}

\title{M-Tuning: Prompt Tuning with Mitigated Label Bias in Open-Set Scenarios}

\author{Ning Liao, Xiaopeng Zhang, Min Cao, Junchi Yan (\textit{IEEE Senior Member}))
\thanks{Ning Liao is with the MoE Key Lab of Artificial Intelligence, Shanghai Jiao Tong University, Shanghai 200240, P. R. China. E-mail: liaoning@sjtu.edu.cn. 

Xiaopeng Zhang is with Huawei Inc., Shenzhen, Guangdong 518129, P. R. China. E-mail: zxphistory@gmail.com.

Min Cao is with the School of Computer Science and Technology, Soochow University, Suzhou 215006, P. R. China. E-mail: caomin0719@126.com.

Junchi Yan is with the School of Artificial Intelligence, Shanghai Jiao Tong University, Shanghai 200240, P. R. China. E-mail: yanjunchi@sjtu.edu.cn. 
\protect\

Correspondence author: Xiaopeng Zhang.\protect
}}

\markboth{IEEE Transactions on Circuits and Systems for Video Technology}%
{How to Use the IEEEtran \LaTeX \ Templates}

\maketitle

\begin{abstract}
In realistic open-set scenarios where labels of a part of testing data are totally unknown, when vision-language (VL) prompt learning methods encounter inputs related to unknown classes (i.e., not seen during training), they always predict them as one of the training classes. The exhibited label bias causes difficulty in open set recognition (OSR), in which an image should be correctly predicted as one of the known classes or the unknown one. To achieve this goal, we propose a vision-language prompt tuning method with mitigated label bias (M-Tuning). It introduces open words from the WordNet to extend the range of words forming the prompt texts from only closed-set label words to more, and thus prompts are tuned in a simulated open-set scenario. Besides, inspired by the observation that classifying directly on large datasets causes a much higher false positive rate than on small datasets, we propose a Combinatorial Tuning and Testing (CTT) strategy for improving performance. CTT decomposes M-Tuning on large datasets as multiple independent group-wise tuning on fewer classes, then makes accurate and comprehensive predictions by selecting the optimal sub-prompt. Finally, given the lack of VL-based OSR baselines in the literature, especially for prompt methods, we contribute new baselines for fair comparisons. Our method achieves the best performance on datasets with various scales, and extensive ablation studies also validate its effectiveness.
\end{abstract}

\begin{IEEEkeywords}
Vision-language, open set recognition, prompt tuning, label bias, combinatorial tuning and testing.
\end{IEEEkeywords}

\section{Introduction}
\IEEEPARstart{P}{rompt} learning on vision-language (VL) pre-trained models~\cite{radford2021learning, jia2021scaling, li2021align, nie2023pro} has shown promising ability in downstream tasks in both data-efficient and parameter-efficient ways. 

For example, CoOp~\cite{zhou2022learning} successfully exploits the CLIP~\cite{radford2021learning} for image recognition by concatenating several learnable prompts with the class names. It shows robustness against the distribution shift from the base dataset, i.e., ImageNet~\cite{russakovsky2015imagenet}, to the target datasets, i.e., the ImageNet variants~\cite{recht2019imagenet, wang2019learning, hendrycks2021natural, hendrycks2021many}, \emph{which have compatible classes with the base dataset}. However, CoOp ignores extending its scalability to new classes out of the limited training set, and has been shown to overfit the base classes, resulting in the poor generalization ability on new classes~\cite{zhou2022conditional}.
\begin{figure}[tb!]
  \centering
  \includegraphics[width=8.5cm]{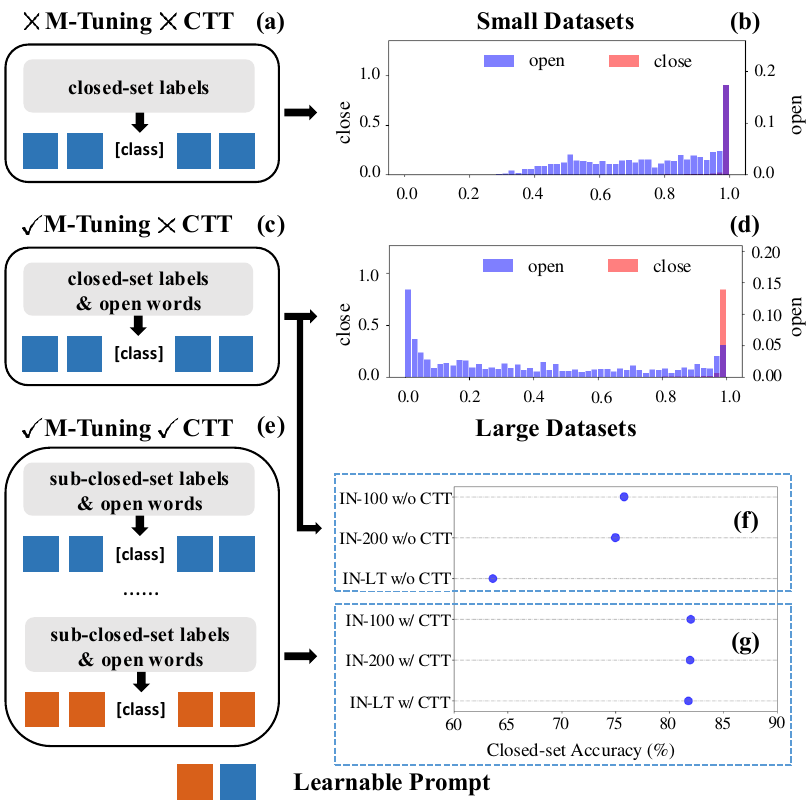}
  \caption{\textbf{Small datasets with fewer classes.} Current prompt tuning methods in (a) only involve known classes, causing the label bias in (b), which shows the distributions of the closed-set maximum probability overlaps significantly between open-set and closed-set data. M-Tuning in (c) introduces open words to extend the range of words forming texts. The results after mitigating the label bias in (d) show the closed-set and open-set data are clearly separated. \textbf{Large datasets with more classes.} In (f) and (g), `IN' is the abbreviation of ImageNet. Prompts represented by different colors are mutual independent. When applying M-Tuning on large datasets directly as in (c), the closed-set accuracy in (f) is low. Combining the CTT strategy in (e), M-Tuning is performed on the decomposed groups with fewer classes, contributing to higher closed-set accuracy in (g).} 
  \label{fig:Fig1}
\end{figure}

To strengthen the generalization ability of prompts from base to new classes, CoCoOp~\cite{zhou2022conditional} integrates the instance information conditioned on each input image into the prompts and achieves great performance under various datasets. In this base-to-new setting~\cite{zhou2022conditional, bulat2022language}, the base classes are utilized for prompt learning in the tuning phase, both base and new classes are filled into prompts in inference. In a word, \emph{the new classes out of the training set are still accessible and known as a priori in the base-to-new generalization.}

However, a more realistic and challenging scenario is that \emph{the part of classes out of the training set are totally unknown to us.} Without knowing new classes, current prompt methods in the Fig.~\ref{fig:Fig1}~(a) suffer from the significant problem in predictions that the distributions of predicted maximum probabilities on base known classes overlap with a great portion between the closed-set known and open-set unknown classes in the Fig.~\ref{fig:Fig1}~(b). Although VL models have been pre-trained on massive data, the label bias, that prompt learning always predict a sample as one of the closed-set classes regardless of whether the class is known, exists objectively. Focusing on mitigating the label bias, we hold that \emph{prompts are expected to equip with the ability of open set recognition (OSR)~\cite{bendale2016towards} to recognize a sample as one of known classes or the unknown one}. The detailed comparisons between zero-shot inference, closed-set classification, distribution shift, base-to-new generalization and open-set recognition are delivered in the Table~\ref{tab:task_diff_comp}.

\begin{table*}[t!]
    \scriptsize
	\centering
	\caption{The detailed comparisons between the prompt-related tasks and representative methods. The tasks include zero-shot inference, closed-set classification, distribution shift, base-to-new generalization and open-set recognition. }
  \label{tab:task_diff_comp}
  \begin{tabular}{ c@{} | c@{} | c@{} | c@{} | c@{}  }
    \toprule
    \midrule
    \,\,Methods\,\, & Tasks &  Descriptions & \,\,\,\makecell{Whether the training and\\ testing classes are compatible?}\,\,\, & \,\,\,\makecell{Whether knowing the true classes\\ out of the training set?}\,\,\, \\
    \midrule
    \,\,\,\makecell{CLIP+ZSL~\cite{radford2021learning}}\,\,\, & Zero-shot inference. & \makecell{Heavily relying on the expertise design.} & N/A & N/A \\
    \midrule
    CoOp~\cite{zhou2022learning} & \multirow{2}{*}{\makecell{Closed-set classification,\\ distribution shift.}} & \multirow{2}{*}{\makecell{Fail in the open world that involves\\ new classes out of the training set.}} & \multirow{2}{*}{Yes} & \multirow{2}{*}{N/A} \\
    UPT~\cite{zang2022unified} & & & &  \\
    \midrule
    CoCoOp~\cite{zhou2022conditional} & \,\,\,\multirow{3}{*}{Base-to-new generalization.}\,\,\, & \multirow{3}{*}{\makecell{Requiring label names of the new\\ classes out of the training set.}} & \multirow{3}{*}{No} & \multirow{3}{*}{Yes} \\
    LASP~\cite{bulat2022language} & & & &  \\
    MAPLE~\cite{khattak2022maple} & & & &  \\
    \midrule
    \,\,\,\textbf{M-Tuning (Ours)}\,\,\, & Open-set recognition. & \,\,\,\makecell{No more information of the testing data out of \\ training classes available, even their names.}\,\,\, & No & No \\
    \bottomrule
  \end{tabular}
\end{table*}

Open-set recognition (OSR)~\cite{junior2017nearest, neal2018open} is a technique for visual recognition in the open world, in which the inference data comes from categories broader than those of training data. It is proposed to mitigate the issue caused by closed-set classifiers that all data will be recognized as the limited training classes. The goal of OSR is to identify whether a test sample belongs to one of the semantic classes in the training phase, and the prediction result of each test sample will be one of the training classes or the unknown. Existing OSR methods~\cite{sun2020conditional, oza2019c2ae, kong2021opengan, zhou2021learning} almost train a simple network from scratch. Recently, the OSR performance has been verified to be highly correlated with the closed-set classifier~\cite{vaze2022open}. In view of the excellent performance achieved by prompt learning in closed-set classification, yet facing the label bias of prompts in realistic classification tasks, we propose the \emph{M-Tuning} to mitigate the label bias of prompt learning in open-set scenarios. Specifically, we focus on the open-set setting based on the pre-trained vision-language model from the downstream tuning-then-testing perspective, regardless of the massive data in the pre-training phase.

The key idea of M-Tuning is to extend the range of words forming the texts from only closed-set classes in the Fig.~\ref{fig:Fig1}~(a) to more. In the distribution shift~\cite{zhou2022learning, zang2022unified} and base-to-new generalization~\cite{zhou2022conditional, bulat2022language}, no matter in the tuning or inference stage, the label words cover exactly all classes. Therefore, the prompts are limited to making predictions from the given classes with high probabilities. To mitigate the label bias, we introduce additional open words from WordNet, that are irrelevant to the training and testing labels, into both the tuning and testing stages in the Fig.~\ref{fig:Fig1}~(c). By doing so, \emph{the prompt learning is put in a simulated open-set scenario}. Each image can be predicted not only from the closed-set classes, but also from the open words. The label bias could thus be mitigated in the Fig.~\ref{fig:Fig1}~(d). The idea and technical details of M-Tuning differ from those of Knowprompt~\cite{chen2022knowprompt}, which aims at incorporating various labels into a comprehensive one by merging the embeddings of multiple words using a weighted sum. M-Tuning is also different from the out-of-distribution detection (OOD) method~\cite{fort2021exploring}, which leverages the true names of outlier classes for outlier exposure. Interestingly, we also prove that it is not necessary to know and use the true labels of open-set data in M-Tuning in our experiments.

In addition, an intuition is that recognition on larger datasets with many classes is more hard than on smaller datasets with few classes. Dividing the large dataset into multiple small ones with fewer classes could decrease the task difficulty. Such an intuition has also been verified in the analysis in MOS~\cite{huang2021mos} that the OOD performance drops significantly with the increase of the in-distribution classes. Also, from the experimental results in the Fig.~\ref{fig:CTT_verify}, using the whole dataset without division ($N_C=1000$) leads to the lowest accuracy. Inspired by the above, we propose the \emph{Combinatorial Tuning and Testing (CTT)} strategy to decompose M-Tuning on large datasets in the Fig.~\ref{fig:Fig1}~(c) as multiple independent group-wise tuning on fewer classes in the Fig.~\ref{fig:Fig1}~(e). In inference, each sample obtains multiple predictions from all the group-specific prompts. The prompt, which exhibits the highest probability within its group-specific closed-set classes, of all prompts is employed as the optimal prediction prompt for a given sample. The closed-set accuracy without CTT in the Fig.~\ref{fig:Fig1}~(f) could be improved by using it in the Fig.~\ref{fig:Fig1}~(g). Different from MOS which devises ``others" for each group and fine-tunes BiT-S~\cite{kolesnikov2020big} by all-data, we introduce the language information and only tune the prompts in a parameter-efficient and data-efficient way.

M-Tuning is the first work of prompt learning in open-set scenarios both on small and large datasets, and thus there is a lack of baselines in the literature. To make fair comparisons, we construct baselines to evaluate the VL-based OSR performance, especially for prompt methods~\cite{radford2021learning, zhou2022learning, zhou2022conditional}. Our method achieves the best performance both on small and large datasets. Extensive ablation experiments validate the effectiveness of each component in our method. 

Our contributions in this paper include:

(1) To mitigate the label bias of prompt learning in OSR, we propose M-Tuning. It extends the range of words forming the texts from only closed-set classes to more, thereby simulating open-set scenarios. To our best knowledge, it is the first vision-language prompt learning method targeting at the challenging open-set scenario.

(2) To effectively deploy the proposed M-Tuning from small to large datasets, we propose the Combinatorial Tuning and Testing (CTT) strategy. It performs the M-Tuning and inference in a divide-then-combine way.

(3) For fair comparisons on VL-based OSR performance, we construct new baselines, especially for prompt methods, which are different from existing OSR methods with training a simple network from scratch.

\section{Related Work}
\subsection{Prompt Learning}
Prompt learning was primarily proposed in natural language processing (NLP)~\cite{petroni2019language, jiang2020can, lester2021power, li2021prefix, liu2021pre, poerner2019bert, shin2020autoprompt} to deploy the pre-trained models with large parameter size in a parameter-efficient and data-efficient way. Owing to its advantages, it has been applied to large-scale vision-language (VL) pre-trained models~\cite{zang2022unified, manas2022mapl, ding2022prompt, li2025toward, xing2023dual, zhu2023prompt, zeng2023temporally, liao2022cohoz, ma2023understanding, li2023cascade, liao2023rethinking} successfully. CoOp~\cite{zhou2022learning} learns continuous task-relevant prompts and exhibits strong robustness to distribution shift on compatible classes. ProDA~\cite{lu2022prompt} is proposed to learn the output embeddings and capture the distributions of prompts towards the target closed-set recognition task. However, the drawback of these methods is the poor generalization ability to classes out of the training set. To learn prompts that can generalize well from base to new classes~\cite{kan2023knowledge, long2023task, zhuBeier2023prompt}, CoCoOp~\cite{zhou2022conditional} integrates instance-specific information into the prompts, which turns the unchanged prompts in CoOp into flexible ones. As the hand-crafted prompts are more generalizable, LASP~\cite{bulat2022language} applies cross entropy loss on texts to close the distance between learned prompts and hand-crafted ones to preserve the generalization ability. To increase the representation capacity of the prompts, LASP additionally introduces the grouping strategy, in which different prompts are optimized towards different set of hand-crafted texts. Similar to LASP~\cite{bulat2022language}, KgCoOP~\cite{yao2023visual} enhances the generalization ability of the learnable prompt by minimizing the discrepancy between the learnable prompts and the handcrafted ones, which are supposed to generalize to new classes well.

\emph{In the base-to-new setting, the new classes are not available in prompt tuning phase, but they are known and directly filled into prompts in testing.} However, in the more realistic scenarios, \emph{we have no knowledge of the new classes}. Instead of equipping the prompts with base-to-new generalization ability in these methods, it is more appropriate to lead the prompts to predict each sample as one of known classes or the unknown one, which is defined as the open set recognition (OSR). Considering the significant label bias exhibited by current prompt methods~\cite{zhou2022learning, zang2022unified, zhou2022conditional, bulat2022language}, we propose the M-Tuning to mitigate the label bias in prompt learning. Besides, the Combinatorial Tuning and Testing (CTT) strategy is devised to deploy M-Tuning from small to large datasets using grouping strategies to improve the performance, and it is different from the grouped LASP~\cite{bulat2022language}. The grouping in CTT is performed based on the data classes and targets at decreasing the task difficulty in large-scale datasets, while grouping in LASP is performed based on the handcrafted texts and targets at richer representations with better generality.

\subsection{Open Set Recognition}
Models trained on limited classes usually suffer from the significant problem that they always predict unknown data in the open world as known classes. To enable the model to recognize unknown classes, open set recognition (OSR) is proposed~\cite{scheirer2012toward, rudd2017extreme, bendale2015towards, miller2021class, sun2020conditional, chen2020learning}. Within the deep learning-based OSR methods, one classical paradigm is the post-processing~\cite{miller2021class, sun2021m2iosr}. CGDL~\cite{sun2020conditional} and C2AE~\cite{oza2019c2ae} take reconstruction errors and latent distributions for unknown recognition in a post-hoc way. Another paradigm is to directly train a classifier including unknown resorting to fake unknown samples~\cite{neal2018open, yu2017open}. \emph{In the OSR community, all the methods are designed by training a simple backbone from scratch}~\cite{neal2018open, lu2022pmal, kong2021opengan, zhou2021learning, zhang2020hybrid, jiang2023openmix}. Recently, by breaking the limit of the backbones, the ability of unknown detection is proved to be highly correlated with the performance of closed-set classification~\cite{vaze2022open}. The observation is valid across datasets, loss functions and backbones. Based on this observation and the excellent closed-set classification performance of prompt learning on VL models, it is rational to apply VL pre-trained model in open-set scenarios, and several methods have been proposed yet. Towards the hierarchical open-granularity classification, the ProTeCt~\cite{wu2023protect} is proposed to calibrate classification across all possible label set granularities using CLIP~\cite{radford2021learning}. Its setting is novel for considering the consistency in the hierarchical predictions, and different from the common OSR setting~\cite{scheirer2012toward, rudd2017extreme, bendale2015towards, miller2021class, sun2020conditional, chen2020learning}. The $A^2PT$~\cite{ren20232} is proposed to solve the OSR problem via prompt learning by its cross-modal guided activation module and anti-association calibration module. However, its experiments follow the all-data setting, and the few-shot experiments in consistent with that of CoOp~\cite{zhou2022learning} and CoCoOp~\cite{zhou2022conditional} have not been performed for comprehensive comparisons. Seeing the lack of baselines for OSR based on VL model, especially for prompt methods, we carefully construct new baselines under all-data and few-shot settings in this paper for fair comparisons.

\subsection{WordNet Applications}
The WordNet  has been widely adopted in various recognition tasks~\cite{shen2022k, bujwid2021large}. For the multiclass classification task, to develop a model allowing for flexible relation encoding, consistent predictions and knowledge transfer, HEX~\cite{bujwid2021large} is proposed by adopting the hierarchical relations in the WordNet for further enhancing the labels, while it is limited to the closed-set classification. Similarly, the hierarchical relation within the WordNet has also been adopted in classification~\cite{chatterjee2023imagenet} together with the proposed soft labels to fully predict the semantics of samples. For the zero-shot classification task solved by prompt learning, handcrafting high performing prompts is prohibitive and difficult, and these prompts are usually not general enough. To solve this issue, CuPL~\cite{pratt2023does} is proposed to augment prompts using large language models such as GPT-3~\cite{brown2020language} or the WordNet. Given the exact candidate class names, more descriptive texts of each class are generated by prompting GPT-3, or taken from the annotation of the class in the WordNet. Such informative texts are then used as prompts for zero-shot classification, instead of using the handcrafted ones. \textit{These methods leverage the WordNet mainly for augmentation on the label hierarchy or the label annotation, based on the premise of knowing the exact classes to be predicted from.} But in the open-set scenarios, we have no knowledge of all classes, thus the above methods are no longer applicable. The goal of utilizing the WordNet in this paper is to expand the label space, which has not been explored for open-set recognition.

\begin{figure*}[tb!]
  \centering
  \includegraphics[width=16cm]{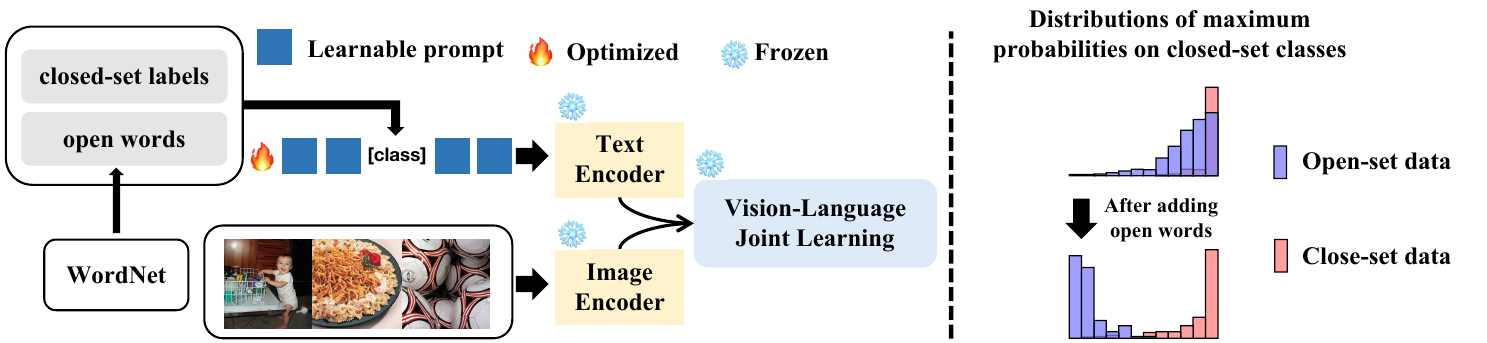}
  \caption{\textbf{Left.} The framework of the proposed M-Tuning. To simulate the open-set scenarios, we introduce open words from WordNet into prompt tuning. The open words are irrelevant to the downstream training and testing classes. Each image could be predicted from the closed-set classes and open words, rather than only the closed-set classes. \textbf{Right.} From the comparison of the distributions of the maximum probabilities on closed-set classes, the label bias of prompt learning is mitigated by M-Tuning.} 
  \label{fig:M_Tuning}
\end{figure*}
\section{Method}
\subsection{Brief Definition of Open Set Recognition}
Referring to the OSR literatures~\cite{scheirer2012toward, bendale2015towards, sun2020conditional}, the OSR task is defined as:

\textit{There are $N$ closed-set known classes given at the training time. After training the model, in testing, $N$ closed-set classes and other $K$ open-set classes exist. The objective of OSR is to correctly classifying the $N$ known classes, as well as recognizing the $K$ open-set classes as a uniform unknown class. Briefly, OSR performs an $N+1$ classification.}

As analyzed in the survey~\cite{salehi2022unified}, the out-of-distribution (OOD) detection problem is canonical to OSR in the multi-class setting, in which the model need to classify the known classes and detect unknown ones. ``\textit{However, OOD detection encompasses a broader spectrum of learning tasks (e.g., multi-label classification, reinforcement learning) and solution space (e.g., density estimation without classification).}"~\cite{salehi2022unified}. 

Therefore, in this paper, we do not extend our scope to OOD detection. Our goal is to learn prompts targeting at the image recognition in open-set scenarios by mitigating the label bias. The setups in experiments and validation are consistent with the OSR literatures~\cite{rudd2017extreme, miller2021class, chen2020learning}.

\subsection{Symbol Notation}
Before introducing our method, we present the symbols and their notations for clear, as delivered in the Table~\ref{tab:syb_note}.
\begin{table}[t!]
    \scriptsize
	\centering
	\caption{Symbol notations.}
  \label{tab:syb_note}
  \begin{tabular}{c@{}|c@{}}
    \toprule
    Symbol \quad\quad & Notation \\
    \midrule
    $\mathcal{D}$ & Dataset \\
    $N_{D}$       & Number of classes in the dataset \\
    $N_O$         & Number of open words \\
    ${E_I}$       & Image encoder \\
    ${E_T}$       & Text encoder \\
    $\boldsymbol{x}$ & Image \\
    $C_i$ & The class of the image \\
    $\boldsymbol{F}()$ & Optimizable prompt \\
    $\boldsymbol{v}$ & Parameterized prompt embedding \\
    $L$ & Number of optimizable embeddings in the prompts \\
    $G$ & Number of groups \\
    $N_C$ & Maximum number of classes within a group \\
    $N_g^k$ & Number of closed-set classes in the $k$-th group \\
    $p_{max}^k$ & Maximum closed-set probability in the $k$-th group \\
    $I_{opt}$ & Index of the optimal prompt \\
    $S_{open}$ & Score of being unknown \\
    \bottomrule
  \end{tabular}
\end{table}

\subsection{M-Tuning}
\label{sec:M_Tuning}
Image recognition promoted by prompt learning almost require the detailed class names both in tuning and inference stages, no matter whether in distribution shift~\cite{zhou2022learning} or base-to-new generalization~\cite{zhou2022conditional, bulat2022language, khattak2022maple}.  The prompts force the images belonging to both known and unknown classes to be predicted as one of the known classes with high probabilities. If the high probabilities can be regularized, by which closed-set classes can still be correctly predicted while open-set unknown images obtain much lower probabilities on known classes, the label bias would be mitigated. 

To this end, we propose the M-Tuning as shown in the Fig.~\ref{fig:M_Tuning}. The underlying rationality is to put the prompt tuning at a simulated open-set scenario, in which each image is predicted not only from the known classes, but also from the open words. To avoid using priors of the open-set classes, the open words are selected after filtering out the class names of the downstream training and testing sets. Specifically, given a dataset denoted as $\mathcal{D}$ with $N_{D}$ classes, we introduce additional $N_O$ open words that exhibit low similarities with the closed-set classes from the WordNet into the prompt tuning phase. The investigation of selecting what kind of open words will be studied in the Sec.~\ref{sec:Further_Investigation}. By doing so, the prompt specific to the closed-set data forms $N_{D}+N_O$ texts that are ready to be matched with each image. Formally, denoting the image encoder and the text encoder in the pre-trained VL model as ${E_I}$ and ${E_T}$, the probability that an image $\boldsymbol{x}$ belonging to its class $C_i$ is calculated on classes indexed within $[0, N_{D} - 1] \cup [N_{D}, N_{D} + N_O - 1]$, and measured by a commonly used cosine metric $<\cdot>$ with the temperature parameter $T$:
\begin{equation}
\begin{aligned}
p(y = {C_i}&|\boldsymbol{x}) = \\
&\frac{{\exp ({{ < {E_I}(\boldsymbol{x}) \cdot {E_T}(\boldsymbol{F}({C_i})) > } \mathord{\left/
 {\vphantom {{ < {E_I}(\boldsymbol{x}) \cdot {E_T}(\boldsymbol{F}({C_i})) > } T}} \right.
 \kern-\nulldelimiterspace} T})}}{{\sum\limits_{j = 0}^{N_{D} + N_O - 1} {\exp ({{ < {E_I}(\boldsymbol{x}) \cdot {E_T}(\boldsymbol{F}({C_j})) > } \mathord{\left/
 {\vphantom {{ < {E_I}(\boldsymbol{x}) \cdot {E_T}(\boldsymbol{F}({C_{i}})) > } T}} \right.
 \kern-\nulldelimiterspace} T})} }},
\label{eq:contras_sim}
\end{aligned}
\end{equation}
in which $\boldsymbol{F}()$ is the optimizable prompt composed of $L$ parameterized embeddings $\boldsymbol{v}$:
\begin{equation}
\begin{aligned}
\boldsymbol{F}(class) = [\boldsymbol{v}]_1 [\boldsymbol{v}]_2 ... [class]  ... [\boldsymbol{v}]_{L}.
\label{eq:template}
\end{aligned}
\end{equation}

The dimension of the parameterized embedding $\boldsymbol{v}$ is the same as the hidden dimension of the image encoder ${E_I}$. We optimize the prompts using cross entropy loss as:
\begin{equation}
\begin{aligned}
L =  - \sum\limits_{\boldsymbol{x} \in \mathcal{D} }{\log (p(y = {C_{i}|\boldsymbol{x}))}}.
\label{eq:contras_loss}
\end{aligned}
\end{equation}

In M-Tuning, only the parameters in prompts are optimized, and the VL model is kept frozen, as shown in the Fig.~\ref{fig:M_Tuning}.

\subsection{Combinatorial Tuning and Testing (CTT)}
\label{sec:CTT}
\begin{figure*}[tb!]
  \centering
  \includegraphics[width=16cm]{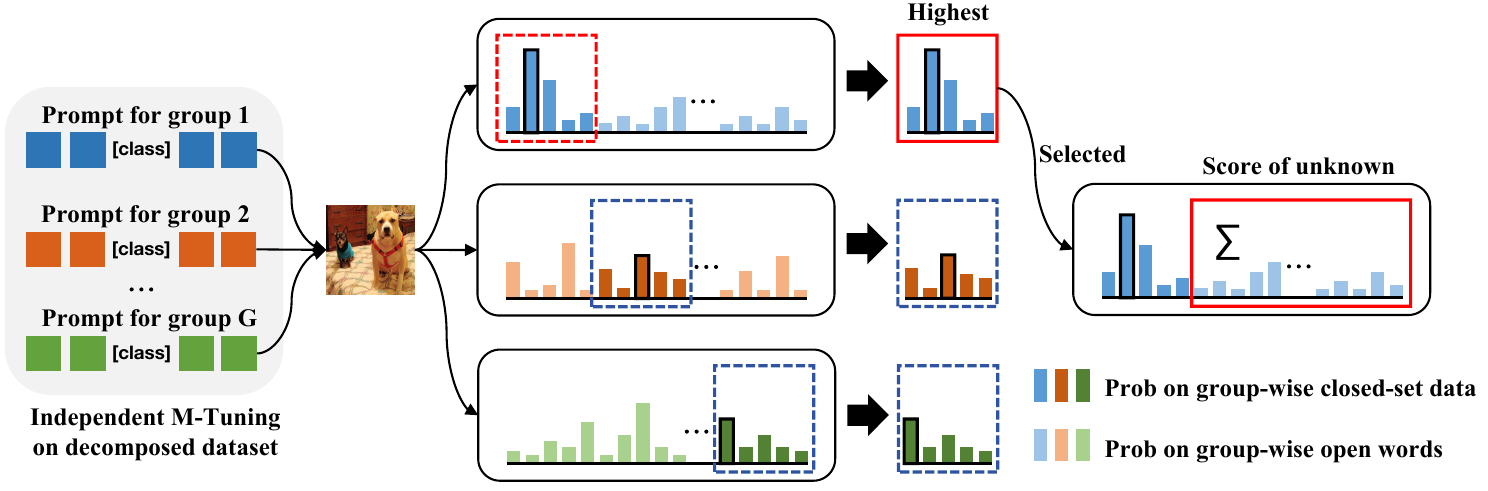}
  \caption{The framework of the proposed Combinatorial Tuning and Testing (CTT) strategy. The datasets are divided into several independent groups by categories. Each group is devised with a set of group-specific prompts. M-Tuning is performed on each group without mutual effect. After tuning, each testing image is predicted by all prompts. By focusing on the group-wise close-set predicted probabilities, the optimal prompt is selected for the final prediction. As a special case, the number of groups is set to 1 for small-scale datasets. } 
  \label{fig:CTT}
\end{figure*}
As demonstrated in MOS~\cite{huang2021mos}, the false positive rate rises with the increasing of in-distribution classes. In order to improve the performance on large datasets with many classes in open-set scenarios, we propose the CTT strategy in the Fig.~\ref{fig:CTT}. It decomposes the tuning and inference from directly on the whole datasets to on the divided datasets with each group containing fewer classes.

\textbf{Dividing Large Datasets into Groups for M-Tuning.} Before M-Tuning, we divide the large datasets consisting of $N_{D}$ known classes into $G$ groups according to the data categories. Therefore, the classes in each group are mutual independent without overlap. The maximum number of classes in each group is $N_C$, and the numbers of classes in the first $G-1$ groups are equal. Formally, we denote the number of classes in each group as $N_g^k,k \in [1,G]$, in which $k$ is the group index. The grouping rule is:
\begin{equation}
\begin{aligned}
(G - 1) \cdot& {N_C} < N_{D} \le G \cdot {N_C},\\
N_g^{G} \le N_g^k &= {N_C},k \in [1,G - 1].
\label{eq:group_rule}
\end{aligned}
\end{equation}

In our main experiments, the grouping is guided by the WordNet ID (WNID) order of class names. After sorting the classes into a list according to the WNID order, the groups are constructed by selecting every adjacent $N_C$ classes in the list continuously. The ablation studies on the grouping strategies are given in Sec.~\ref{sec:grouping_ablation}. 

To ensure the M-Tuning on each group is mutually independent, as shown in the Fig.~\ref{fig:CTT} left part, we devise a set of group-specific prompts for each group:
\begin{equation}
\begin{aligned}
\boldsymbol{F}_{k}(class) = [\boldsymbol{v}]_1^{k} [\boldsymbol{v}]_2^{k} ... [class]  ... [\boldsymbol{v}]_{L}^{k}.
\label{eq:template_group}
\end{aligned}
\end{equation}

For each group, the prompt texts consist of $N_g^k$ group-wise closed-set classes and $N_O$ open words. Therefore, in accordance with Eq.~\ref{eq:contras_sim}, the group-wise predicted probability that an image $\boldsymbol{x}$ belonging to its class $C_i^k$ is calculated on group-wise classes indexed within $[0, N_g^k - 1] \cup [N_g^k, N_g^k + N_O - 1], k \in [1, G]$ as:
\begin{equation}
\begin{aligned}
p(y = {C_{i}^{k}}&|\boldsymbol{x}) = \\
&\frac{{\exp ({{ < {E_I}(\boldsymbol{x}) \cdot {E_T}(\boldsymbol{F}_k({C_{i}^{k}})) > } \mathord{\left/
 {\vphantom {{ < {E_I}(\boldsymbol{x}) \cdot {E_T}(\boldsymbol{F}({C_{i}^{k}})) > } T}} \right.
 \kern-\nulldelimiterspace} T})}}{{\sum\limits_{j = 0}^{N_g^k + N_O - 1} {\exp ({{ < {E_I}(\boldsymbol{x}) \cdot {E_T}(\boldsymbol{F}_k({C_{j}^{k}})) > } \mathord{\left/
 {\vphantom {{ < {E_I}(\boldsymbol{x}) \cdot {E_T}(\boldsymbol{F}_k({C_{i}})) > } T}} \right.
 \kern-\nulldelimiterspace} T})} }},
\label{eq:contras_sim_group}
\end{aligned}
\end{equation}

When M-Tuning is performed on a group with its corresponding prompts being optimized by the loss function defined in Eq.~\ref{eq:contras_loss}, the prompts of other groups are kept frozen and not utilized in the forward process to prevent their mutual effect. 

\textbf{Combinatorial Testing.} To perform a comprehensive predictions combining all group-specific ones, we propose the combinatorial inference as shown in the Fig.~\ref{fig:CTT} right part, in which open words are also introduced to keep the label bias being mitigated. Regardless of whether the testing image belongs to a known class or an unknown one, each testing image will be predicted by all group-specific prompts. Specifically, each group-wise prompt is filled with $N_g^k, k \in [1, G]$ group-wise closed-set classes and $N_O$ open words, and forming $N_g^k + N_O, k \in [1, G]$ texts as the input of the text encoder. 

In the prediction of a group-specific prompt, only the group-specific closed-set probabilities carry actual meanings of how likely the test sample belongs to these classes, as shown from the dashed rectangles in the Fig.~\ref{fig:CTT}. Probabilities on group-wise open words are worthless for selecting the optimal sub-prompt for final predictions. Therefore, we define the group-specific closed-set maximum probability $p_{max}^k, k \in [1,G]$ as:
\begin{equation}
\begin{aligned}
{p_{max}^k} = \max (p(y = {C_j^{k}}|\boldsymbol{x})),k \in [1,G], j \in [0, N_g^k-1].
\label{eq:max_close}
\end{aligned}
\end{equation}

As a test sample will be predicted with a high probability on its true class by the prompt corresponding to its group, we choose the optimal prompt with the group indexed as $I_{opt}$:
\begin{equation}
\begin{aligned}
{I_{opt}} = \arg\max {p_{max}^k},k \in [1,G].
\label{eq:optimal_id}
\end{aligned}
\end{equation}

\subsection{Evaluation Indicators}
\textbf{For binary unknown detection.} Considering the sum of probabilities on group-specific closed-set classes to be the score of being known, other probabilities are therefore taken for binary detection by a defined score $S_{open}$ denoting a sample being unknown as:
\begin{equation}
\begin{aligned}
{S_{open}} = 1 - \sum\limits_{j = 0}^{{N_g^{I_{opt}}} - 1} {p(y = {C_j^{I_{opt}}}|\boldsymbol{x})},
\label{eq:open_score}
\end{aligned}
\end{equation}
in which $N_g^{I_{opt}}$ is the number of closed-set classes in the selected optimal group indexed as $I_{opt}$ in Eq.~\ref{eq:optimal_id}, $C_j^{I_{opt}}$ is the $j$-th class in the group corresponding to the optimal prompt.

\textbf{For detailed open-set recognition.} After selecting the optimal prompt indexed by $I_{opt}$ for a sample, the detailed prediction in this group is made by setting a threshold $\tau_{max}$ on the group-specific closed-set maximum probability $p_{max}^{I_{opt}}$. If $p_{max}^{I_{opt}}$ is no less than $\tau_{max}$, which means the prediction is of high confidence, the sample is predicted as one of the closed-set classes in this group ranging from 0 to $N_g^{I_{opt}}-1$ using argmax; otherwise, the sample is predicted as unknown. Formally, the prediction is specified as:
\begin{equation}
\begin{aligned}
pred = \left\{ \begin{array}{l}
\mathop {\arg \max }\limits_{j \in [0,{N_g^{I_{opt}}} - 1]} p(y = {C_j^{I_{opt}}}|\boldsymbol{x}), \,\, if \,\, {p_{max}^{I_{opt}}} \ge \tau_{max} \\
{\rm \,\,\,\,\,unknown},else
\end{array} \right..
\label{eq:thresh_pred}
\end{aligned}
\end{equation}

CTT is not only applicable on large datasets. Specifically, the number of groups is 1 for small datasets.

\section{Experiments}
\subsection{Implementation Details}
In our experiments, we use the pre-trained contrastive language-image pre-training (CLIP)~\cite{radford2021learning} as the backbone, in which the image encoder is the base version ViT-B/16~\cite{dosovitskiy2020image}. In consistent with CoOp~\cite{zhou2022learning} and CoCoOp~\cite{zhou2022conditional}, the learnable prompts are randomly initialized by drawing from a Gaussian distribution with mean value equal to 0 and standard deviation equal to 0.02. The $[class]$ is placed in the middle of prompts, whose length $L$ is set as 10. The ablation study of the prompt format is given in Sec.~\ref{sec:ablation_prompt_formats}. In the M-Tuning phase, the initial learning rate is set to $1e-5$. We apply the linear learning rate decay scheduler to the AdamW optimizer~\cite{loshchilov2018decoupled} as suggested by the default setup of Huggingface Transformers~\footnote{https://huggingface.co/transformers/.}. The temperature parameter $T$ is set to be consistent with that in CLIP, without any scale. M-Tuning is performed for 30 epochs on 4 NVIDIA Tesla V100 GPUs with batch size 64. For fair comparisons with the constructed baselines in Sec.~\ref{sec:baseline_methods}, we perform the main experiments of M-Tuning both under the all-data setting and the few-shot setting, in which 16 shots per class are selected. 

As for the datasets, because the open-set setting is quite different from the setting of zero-shot prediction, distribution shift and base-to-new generalization in existing CoOp-based methods~\cite{zhou2022learning, zhou2022conditional} as shown in the Table~\ref{tab:task_diff_comp}, the datasets adopted in these methods are not suitable for experiments in open-set scenarios. In addition, there already exists datasets that are widely used for experiments in previous open-set recognition methods~\cite{neal2018open, lu2022pmal, kong2021opengan, zhou2021learning, zhang2020hybrid, jiang2023openmix}, thus we keep using these datasets, which will be introduced in detail as below.

\subsection{Evaluation Metrics}
Following OSR protocols~\cite{neal2018open, zhou2021learning}, there are 3 metrics adopted for performance evaluation:

\textbf{AUROC.} For unknown detection task, which is a binary recognition in predicting whether a test sample is known or unknown, the AUROC (Area Under ROC Curve) is adopted. It is a calibration-free measure that could comprehensively characterize the performance for a given score, i.e., the $S_{open}$ defined in Eq.~\ref{eq:open_score}, by varying thresholds. 

\textbf{Accuracy.} We additionally use accuracy to measure the closed-set classification performance in unknown detection.

\textbf{mF1-score.} To measure the performance of predicting each sample as one of the known classes or the unknown one as depicted in Eq.~\ref{eq:thresh_pred} in detail, the mF1-score (macro-averaged F1-score) is used for overall evaluation.

\subsection{Baseline Methods}
\label{sec:baseline_methods}
Given the lack of baselines of VL-based OSR, especially for prompt learning methods, we carefully construct new baselines for fair comparisons with other prompt learning and OSR methods. They are:

\textbf{CLIP+ZSL~\cite{radford2021learning}.} Without prompt tuning the CLIP model with the image encoder selected as the ViT-B/16~\cite{dosovitskiy2020image}, we use the released 80 handcrafted templates~\footnote{https://github.com/openai/CLIP.} for zero-shot learning (ZSL) prediction. Besides, for performance evaluation, we use the maximum softmax probability (MSP)~\cite{hendrycks2016baseline} as the indicator, which has been commonly used~\cite{neal2018open, zhang2020hybrid}.

\textbf{CoOp~\cite{zhou2022learning} and CoCoOp~\cite{zhou2022conditional}.} As reported in the paper~\cite{zhou2022learning, zhou2022conditional}, the experiments of these two methods are all performed in the 16-shot setting. Following this setting, we implement CoOp and CoCoOp by selecting 16 shots per class. Specifically, the architecture of the vision model in the CLIP is selected as the ViT-B/16~\cite{dosovitskiy2020image}. Different from the base-to-new generalization setting, the true names of the open-set data are not available. Therefore, only the closed-set labels are accessible both in the tuning and testing phases. The prompts are optimized toward maximizing the similarities between images and closed-set category-filled texts. We use the same training hyperparameters in the publicly released code~\footnote{https://github.com/KaiyangZhou/CoOp.} for building this baseline. MSP~\cite{hendrycks2016baseline} is adopted here for performance evaluation. 

\textbf{LASP~\cite{bulat2022language} and KgCoOp~\cite{yao2023visual}.} These two methods are originally proposed for base-to-new generalization by adding a constraint between the learnable prompts and the handcrafted ones with great generalization ability. Under the 16-shot setting, we re-implement them without knowing the true label names of the open-set data. We use the version of CLIP with the vision model selected as the ViT-B/16~\cite{dosovitskiy2020image}. For LASP~\cite{bulat2022language}, the textual prompt used to maintain the generalization ability of the learnable one is designed as ``\textit{a photo of \{\}}". In addition to the text-image similarity, the similarity between the textual and the learnable prompt is also counted in the optimization process. We keep using the same hyperparameters in the public code~\footnote{https://github.com/1adrianb/lasp.} of LASP for building this baseline. For KgCoOp~\cite{yao2023visual}, the handcrafted prompt is designed as ``\textit{a photo of a \{\}}", which is devised as another supervision from the perspective of generalization ability, instead of only the text-image similarity. This baseline is implemented by keeping the hyperparameters unchanged in the public code~\footnote{https://github.com/htyao89/KgCoOp.} of KgCoOp. The performance evaluation is conducted by leveraging the MSP~\cite{hendrycks2016baseline} as the indicator.

\textbf{ARPL~\cite{chen2021adversarial}.} As a commonly compared state-of-the-art OSR method, ARPL is implemented on the image encoder, i.e., ViT-B/16, of CLIP for comparison. The above baselines and our method only optimize a few parameters with the backbone kept frozen, thus the ARPL is also implemented with the image encoder of CLIP kept frozen following its official all-data setting~\cite{chen2021adversarial}. This baseline is constructed by using the released code~\footnote{https://github.com/iCGY96/ARPL.} and the hyperparameters with the same training objectives.

\textbf{Cross-Entropy+ (CE+)~\cite{vaze2022open}.} CE+ trains the classifier by cross entropy simply using the maximum logit rather than the softmax one. This baseline is implemented using the official code~\footnote{https://github.com/sgvaze/osr\_closed\_set\_all\_you\_need.} with the same hyperparameters and training objectives on the image encoder, i.e., ViT-B/16, of CLIP.  Only the classification head is optimized following its all-data setting.

Apart from the above baselines, other OSR methods are not considered for a fair comparison, since they require additional parameters to be optimized in the encoder-decoder structure~\cite{oza2019c2ae, sun2020conditional, guo2021conditional, yue2021counterfactual} or the generator-discriminator structure~\cite{kong2021opengan, neal2018open, moon2022difficulty}.
\begin{table*}[t!]
    \scriptsize
	\centering
	\caption{Performance comparisons with baselines in the unknown detection on small-scale datasets. The results of the compared methods (CLIP+ZSL~\cite{radford2021learning}, CoOp~\cite{zhou2022learning}, CoCoOp~\cite{zhou2022conditional}, LASP~\cite{bulat2022language}, KgCoOp~\cite{yao2023visual}, ARPL~\cite{chen2021adversarial}, CE+~\cite{vaze2022open}) are all self-implemented to build a new baseline for the lack of baselines of VL-based OSR performance.}
  \label{tab:unknown_detection_smalldata}
  \begin{tabular}{c@{} |c@{} c@{} c@{} c@{} c@{} | c@{} c@{} c@{} c@{} c@{}}
    \toprule
    \midrule
    \multirow{2}{*}{\,\,Methods\,\,\,\,\,\,} &  \multicolumn{5}{c|}{AUROC} &  \multicolumn{5}{c}{Closed-set Accuracy} \\
    \cline{2-11}
                            & \,\,\,CIFAR10\,\,\, & \,\,\,CIFAR+10\,\,\, & \,\,\,CIFAR+50\,\,\, & \,\,\,TinyImageNet\,\,\, & \,\,\,CIFAR100\,\,\,\,\, & \,\,\,CIFAR10\,\,\, & \,\,\,CIFAR+10\,\,\, & \,\,\,CIFAR+50\,\,\, & \,\,\,TinyImageNet\,\,\, & \,\,\,CIFAR100\,\,\,  \\
    \midrule
    CLIP+ZSL~\cite{radford2021learning}\,\,\, & 81.09 & 82.69 & 86.96 & 82.58 & 86.08 & 84.93 & 88.20 & 88.20 & 86.60 & 83.20 \\
    \midrule
    CoOp~\cite{zhou2022learning} (16-shot)              & 81.22 & 86.46 & 86.39 & 83.08 & 85.59 & 80.10 & 88.45 & 88.20 & 87.00 & 86.25 \\
    CoCoOp~\cite{zhou2022conditional} (16-shot)\,\,         & 84.84 & 85.53 & 87.07 & 85.19 & 85.29 & 81.75 & 89.65 & 89.48 & \textbf{90.00} & 87.80 \\
    LASP~\cite{bulat2022language} (16-shot) & 79.38 & 84.19 & 84.73 & 84.46 & 85.52 & 77.77 & 87.40 & 87.40 & 88.40 & 82.40 \\
    KgCoOp~\cite{yao2023visual} (16-shot) & 82.62 & 86.65 & 87.10 & 85.35 & 86.92 & 83.02 & 87.00 & 89.18 & 88.90 & 87.40 \\
    \textbf{Ours (16-shot)}                             & \textbf{94.66} & \textbf{94.67} & \textbf{95.03} & \textbf{85.85} & \textbf{87.38} & \textbf{92.58} & \textbf{93.73} & \textbf{92.78} & 86.60 & \textbf{87.95} \\
    \midrule
    ARPL~\cite{chen2021adversarial} (all-data)          & 83.86 & 90.76 & 89.12 & 83.24 & 90.68 & 88.18 & 88.65 & 91.33 & 89.70 & 94.00 \\
    CE+~\cite{vaze2022open} (all-data)       & 83.89 & 91.03 & 89.30 & 83.75 & \textbf{90.75} & 88.23 & 88.70 & 91.35 & 90.70 & \textbf{94.05} \\
    \textbf{Ours (all-data)}                            & \textbf{96.29} & \textbf{96.28} & \textbf{96.17} & \textbf{87.30} & 89.51 & \textbf{96.30} & \textbf{96.20} & \textbf{96.23} & \textbf{90.70} & 88.45     
   \\
    \bottomrule
  \end{tabular}
\end{table*}
\begin{table*}[t!]
    \scriptsize
	\centering
	\caption{Performance comparisons with baselines in the unknown detection on large-scale datasets. The results of the compared methods (CLIP+ZSL~\cite{radford2021learning}, CoOp~\cite{zhou2022learning}, CoCoOp~\cite{zhou2022conditional}, LASP~\cite{bulat2022language}, KgCoOp~\cite{yao2023visual}, ARPL~\cite{chen2021adversarial}, CE+~\cite{vaze2022open}) are all self-implemented to build a new baseline for the lack of baselines of VL-based OSR performance.}
  \label{tab:unknown_detection_largedata}
  \begin{tabular}{c@{} |c@{} c@{} c@{}|c@{} c@{} c@{}}
    \toprule
    \midrule
    \multirow{2}{*}{\,\,Methods\,\,\,\,\,\,} &  \multicolumn{3}{c|}{AUROC} &  \multicolumn{3}{c}{Closed-set Accuracy} \\
    \cline{2-7}
                            & \,\,\,ImageNet-100\,\,\, & \,\,\,ImageNet-200\,\,\, & \,\,\,ImageNet-LT\,\,\, & \,\,\,ImageNet-100\,\,\, & \,\,\,ImageNet-200\,\,\, & \,\,\,ImageNet-LT\,\,\, \\
    \midrule
    CLIP+ZSL~\cite{radford2021learning} \,\,\, & 88.93 & 87.49 & 71.99 & 70.92 & 71.48 & 63.82 \\
    \midrule
    CoOp~\cite{zhou2022learning} (16-shot)              & 94.61 & 92.40 & 76.71 & 64.80 & 70.86 & 67.32 \\
    CoCoOp~\cite{zhou2022conditional} (16-shot)\,\,        & 93.54 & 90.54 & 75.63 & 67.44 & 74.25 & 67.29 \\
    LASP~\cite{bulat2022language} (16-shot) & 89.12 & 90.95 & 75.58 & 74.14 & 73.41 & 67.34 \\
    KgCoOp~\cite{yao2023visual} (16-shot) & 86.72 & 88.25 & 76.68 & 73.76 & 73.45 & 67.96 \\
    \textbf{Ours (16-shot)}                             & \textbf{99.01} & \textbf{97.27} & \textbf{81.51} & \textbf{74.34} & \textbf{76.62} & \textbf{83.26}  \\
    \midrule
    ARPL~\cite{chen2021adversarial} (all-data)          & 97.81 & 96.16 & 62.55 & 80.50 & 79.02 & 16.73 \\
    CE+~\cite{vaze2022open} (all-data)                  & 97.65 & 95.71 & 59.18 & 76.68 & 71.30 & 9.79 \\
    \textbf{Ours (all-data)}                            & \textbf{98.93} & \textbf{96.47} & \textbf{78.89} & \textbf{81.96} & \textbf{81.90} & \textbf{81.74} \\
    \bottomrule
  \end{tabular}
\end{table*}
\begin{figure*}[tb!]
\begin{minipage}[b]{4cm}
  \centering
  \centerline{\includegraphics[width=4.5cm]{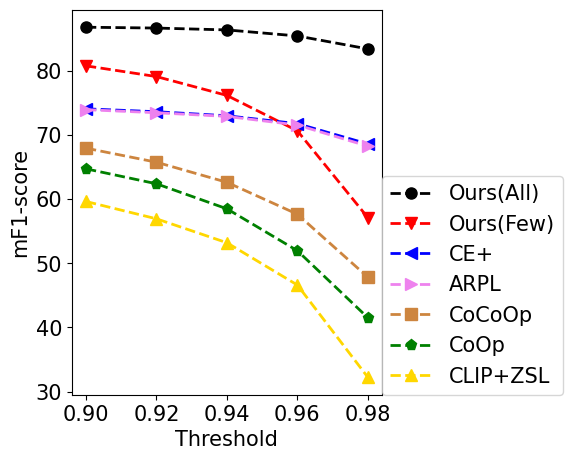}}
  \centerline{\footnotesize (a) ImageNet-crop as the open-set.}
\end{minipage}
\hfill
\begin{minipage}[b]{4cm}
  \centering
  \centerline{\includegraphics[width=4.5cm]{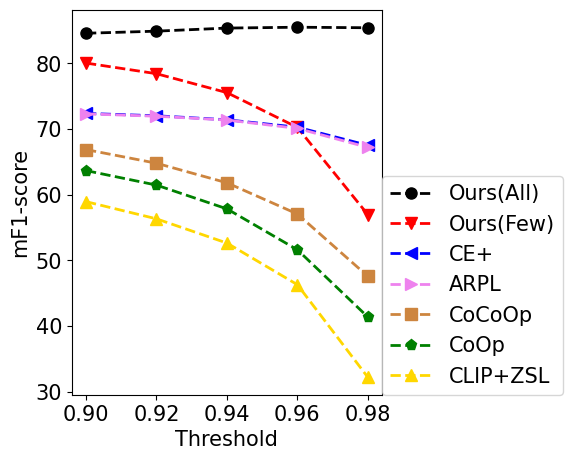}}
  \centerline{\footnotesize (b) ImageNet-resize as the open-set.}
\end{minipage}
\hfill
\begin{minipage}[b]{4cm}
  \centering
  \centerline{\includegraphics[width=4.5cm]{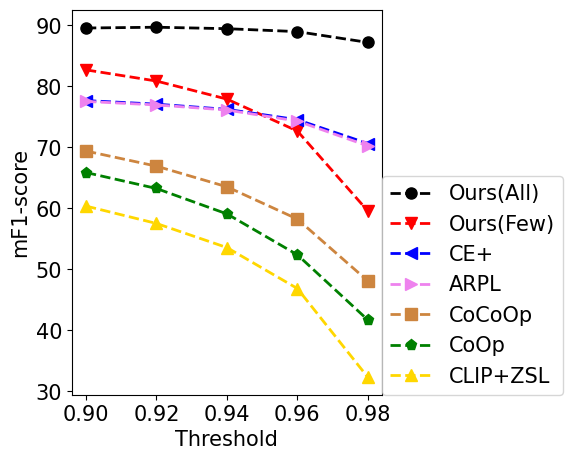}}
  \centerline{\footnotesize (c) LSUN-crop as the open-set. }
\end{minipage}
\hfill
\begin{minipage}[b]{4cm}
  \centering
  \centerline{\includegraphics[width=4.5cm]{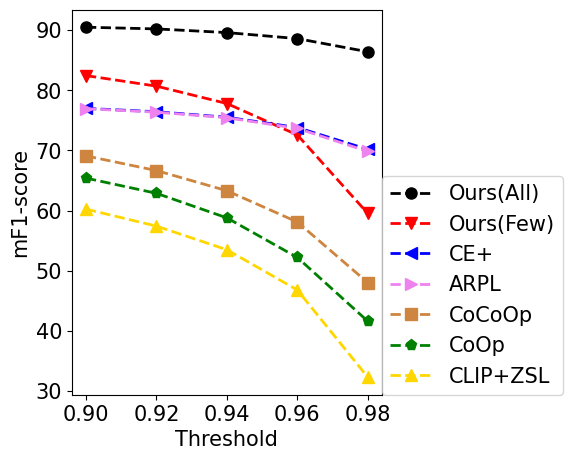}}
  \centerline{\footnotesize (d) LSUN-resize as the open-set. }
\end{minipage}
\hfill
\caption{Comparisons of the detailed recognition results measured by mF1-score. Ours(All) and Ours(Few) denote the results achieved by our method under the all-data and 16-shot settings, respectively. The results of the compared methods (CLIP+ZSL~\cite{radford2021learning}, CoOp~\cite{zhou2022learning}, CoCoOp~\cite{zhou2022conditional}, ARPL~\cite{chen2021adversarial}, CE+~\cite{vaze2022open}) are all self-implemented to build a new baseline for the lack of baselines of VL-based OSR performance.} 
\label{fig:mF1_score}
\end{figure*}

\subsection{Comparisons with Baseline Methods}
\textbf{Unknown Detection on Small Datasets.} The small datasets adopted in the unknown detection experiment and their settings are as follows. CIFAR10~\cite{krizhevsky2009learning} is split into 6 known and 4 unknown classes. The 100 classes in CIFAR100~\cite{krizhevsky2009learning} are divided into 20 known and 80 unknown classes. For CIFAR+10 (resp. +50)~\cite{neal2018open}, 4 classes are selected from CIFAR10 as known, 10 (resp. +50) classes are sampled from CIFAR100 as unknown. TinyImageNet~\cite{le2015tiny} includes 200 classes, of which 20 are known and the remaining 180 are unknown.

For fair comparisons with the baselines, our method is carried out under the 16-shot and the all-data settings. Specifically, the group number $G$ of the proposed CTT is 1 in inference. The unknown detection performance evaluated by AUROC and the closed-set classification performance evaluated by accuracy are given in the Table~\ref{tab:unknown_detection_smalldata}.

\emph{Comparison with CLIP-based prompt methods.} From Table~\ref{tab:unknown_detection_smalldata}, our method surpasses CLIP+ZSL, CoOp, CoCoOp, LASP and KgCoOp by a large margin. Specifically, though CoCoOp generalizes well from base to new classes, without knowing the true names of open-set data, it still fails in open-set scenarios, in which we have no knowledge of the unseen classes, even their names. LASP and KgCoOp are both designed for improving the generalization ability of the prompts by adding a constraint between the learnable prompts and handcrafted ones, which are supposed to be equipped with great generalization ability. However, these two methods still rely on knowing the true label names of the testing data out of training classes, thus perform not as well as ours in the open-set scenarios. The comparisons prove the effectiveness of our method in mitigating the label bias of the prompts, contributing to the great performance in unknown detection on small-scale datasets. 

\emph{Comparison with OSR methods.} Our method also outperforms the OSR methods, i.e., ARPL and CE+, which are implemented on the image encoder of CLIP with the backbone kept frozen. The comparisons reveal the advantages of prompt learning in utilizing the pre-trained knowledge for downstream tasks in a parameter-efficient way. Specifically, by mitigating the label bias of prompt learning, M-Tuning is effective in open-set scenarios.

\textbf{Unknown Detection on Large Datasets.} Three large datasets are used for comparison in unknown detection. Following the dataset preparation~\cite{yang2020convolutional, chen2020learning, lu2022pmal} on the ImageNet dataset with 1000 classes in total, we build ImageNet-100 by selecting the first 100 classes as known and the remaining 900 ones as unknown, and build ImageNet-200 with the first 200 classes as known and 800 ones as unknown. ImageNet-LT~\cite{liu2019large}, as a long-tailed dataset, includes 1000 known classes from ImageNet-2012~\cite{russakovsky2015imagenet}, and 360 unknown classes from the validation set of ImageNet-2010. The number of images in known classes ranges from 5 to 1280. We conduct experiments by dividing the closed-set dataset using the WNID order and setting the maximum number of classes in each group $N_C$ as 20. The experimental results are given in the Table~\ref{tab:unknown_detection_largedata}.

\emph{Comparison with CLIP-based prompt methods.} Compared with the CLIP+ZSL and CoOp, our method performs better. Concerning the methods including CoCoOp, LASP, and KgCoOP, they are designed for equipping the prompts with enhanced generalization ability, while still exhibit performance weakness compared with our method. On the one hand, the results show the effectiveness of the M-Tuning in mitigating the label bias of the prompts in open-set scenarios. On the other hand, the CLIP-based prompt methods only learn a single set of prompt, while our method decreases the task difficulty in a divide-then-combine way by means of the CTT strategy, which verifies the effectiveness of the CTT strategy.
 
\emph{Comparison with OSR methods.} An unsatisfactory performance on the ImageNet-LT is given by the ARPL and CE+. \textit{They are designed for small-scale datasets with balanced numbers of samples per class, and thus may not be applicable to the ImageNet-LT, in which the numbers of samples in each class different from each other significantly.} It is also notable that the closed-set accuracy of our method under the 16-shot setting is better than that under the all-data setting on the ImageNet-LT dataset, we guess that the few-shot setting mitigates the negative effects caused by the imbalanced distributions of class numbers. Table~\ref{tab:unknown_detection_largedata} verifies the wide applicability and effectiveness achieved by CTT.
\begin{figure*}[t!]
\begin{minipage}[b]{4cm}
  \centering
  \centerline{\includegraphics[width=4cm]{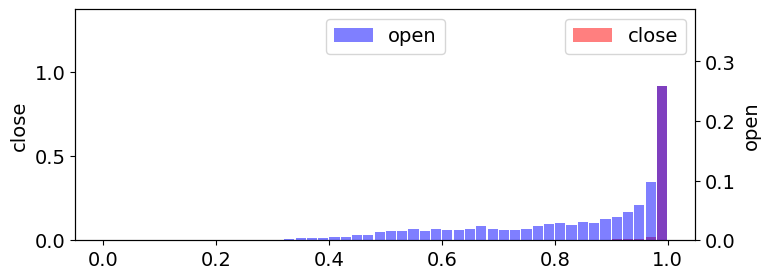}}
  \centerline{\footnotesize (a) CIFAR10, $N_O=0$. }
\end{minipage}
\hfill
\begin{minipage}[b]{4cm}
  \centering
  \centerline{\includegraphics[width=4cm]{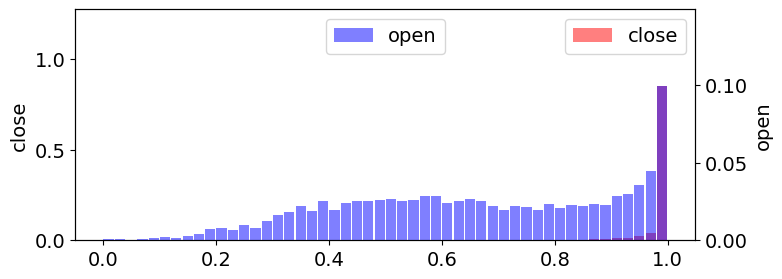}}
  \centerline{\footnotesize (b) CIFAR10, $N_O=20$. }
\end{minipage}
\hfill
\begin{minipage}[b]{4cm}
  \centering
  \centerline{\includegraphics[width=4cm]{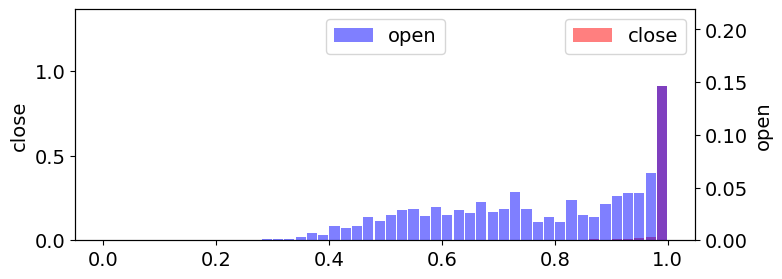}}
  \centerline{\footnotesize (c) CIFAR+10, $N_O=0$. }
\end{minipage}
\hfill
\begin{minipage}[b]{4cm}
  \centering
  \centerline{\includegraphics[width=4cm]{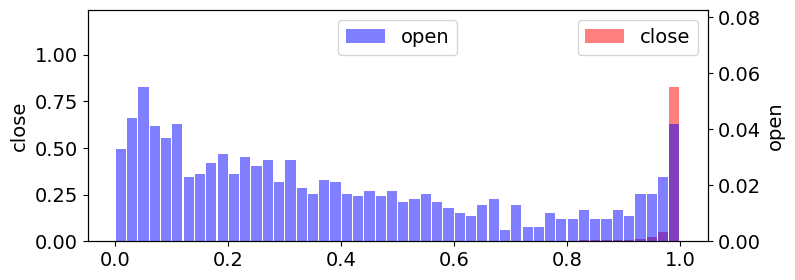}}
  \centerline{\footnotesize (d) CIFAR+10, $N_O=20$. }
\end{minipage}

\begin{minipage}[b]{4cm}
  \centering
  \centerline{\includegraphics[width=4cm]{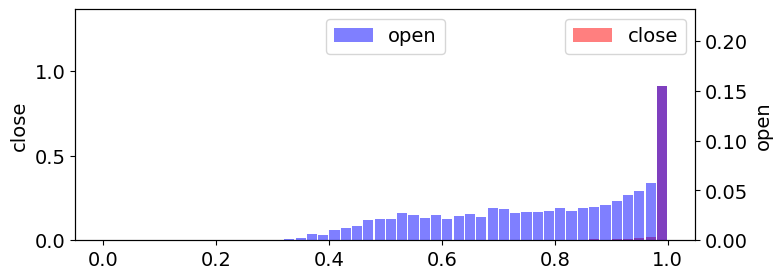}}
  \centerline{\footnotesize (e) CIFAR+50, $N_O=0$. }
\end{minipage}
\hfill
\begin{minipage}[b]{4cm}
  \centering
  \centerline{\includegraphics[width=4cm]{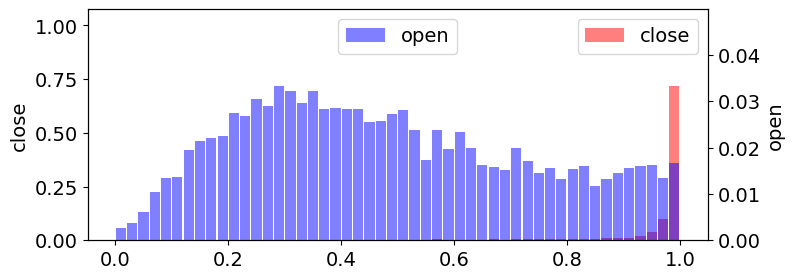}}
  \centerline{\footnotesize (f) CIFAR+50, $N_O=10$. }
\end{minipage}
\hfill
\begin{minipage}[b]{4cm}
  \centering
  \centerline{\includegraphics[width=4cm]{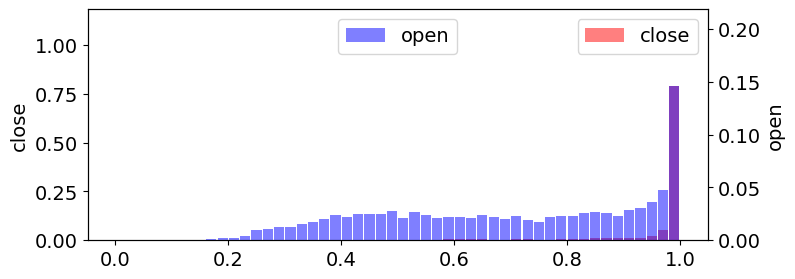}}
  \centerline{\footnotesize (g) TinyImageNet, $N_O=0$. }
\end{minipage}
\hfill
\begin{minipage}[b]{4cm}
  \centering
  \centerline{\includegraphics[width=4cm]{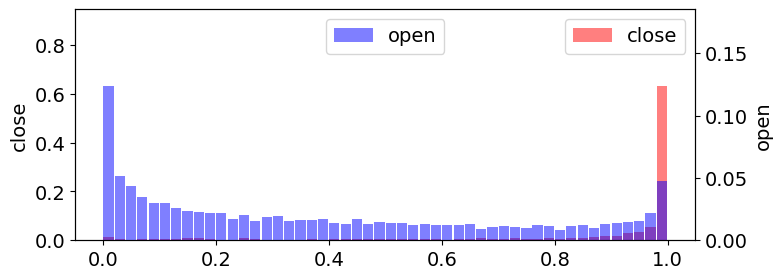}}
  \centerline{\footnotesize (h) TinyImageNet, $N_O=20$. }
\end{minipage}

\begin{minipage}[b]{4cm}
  \centering
  \centerline{\includegraphics[width=4cm]{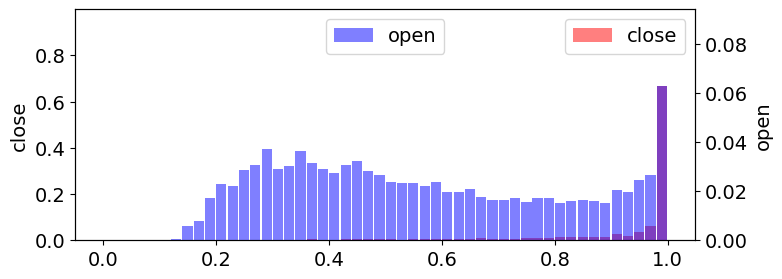}}
  \centerline{\footnotesize (i) CIFAR100, $N_O=0$. }
\end{minipage}
\hfill
\begin{minipage}[b]{4cm}
  \centering
  \centerline{\includegraphics[width=4cm]{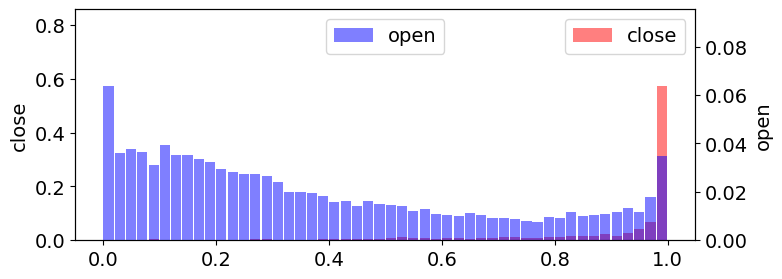}}
  \centerline{\footnotesize (j) CIFAR100, $N_O=20$. }
\end{minipage}
\hfill
\begin{minipage}[b]{4cm}
  \centering
  \centerline{\includegraphics[width=4cm]{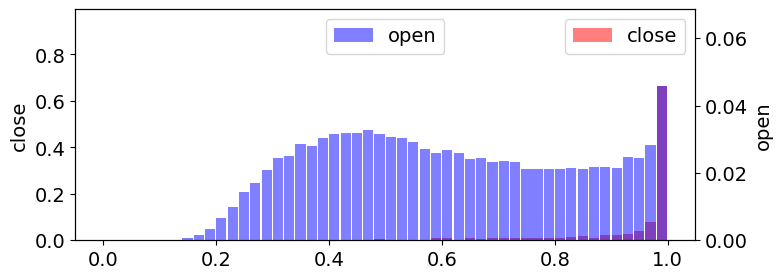}}
  \centerline{\footnotesize (k) ImageNet-100, Group 0, $N_O=0$. }
\end{minipage}
\hfill
\begin{minipage}[b]{4cm}
  \centering
  \centerline{\includegraphics[width=4cm]{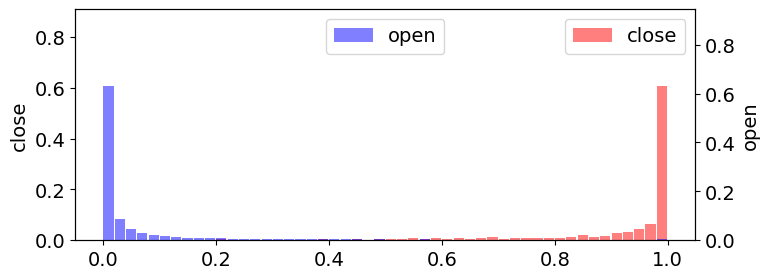}}
  \centerline{\footnotesize (l) ImageNet-100, Group 0, $N_O=60$. }
\end{minipage}

\begin{minipage}[b]{4cm}
  \centering
  \centerline{\includegraphics[width=4cm]{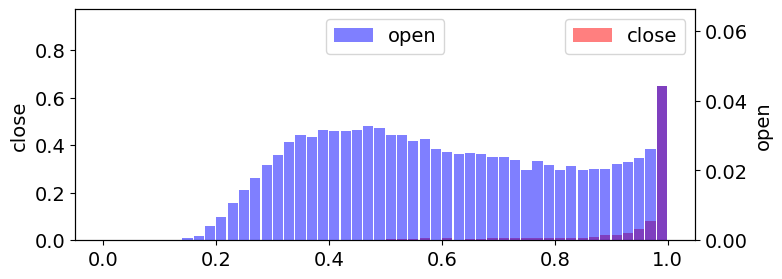}}
  \centerline{\footnotesize (m) ImageNet-200, Group 0, $N_O=0$. }
\end{minipage}
\hfill
\begin{minipage}[b]{4cm}
  \centering
  \centerline{\includegraphics[width=4cm]{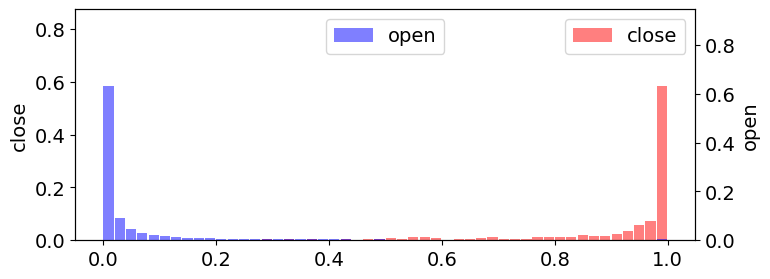}}
  \centerline{\footnotesize (n) ImageNet-200, Group 0, $N_O=60$. }
\end{minipage}
\hfill
\begin{minipage}[b]{4cm}
  \centering
  \centerline{\includegraphics[width=4cm]{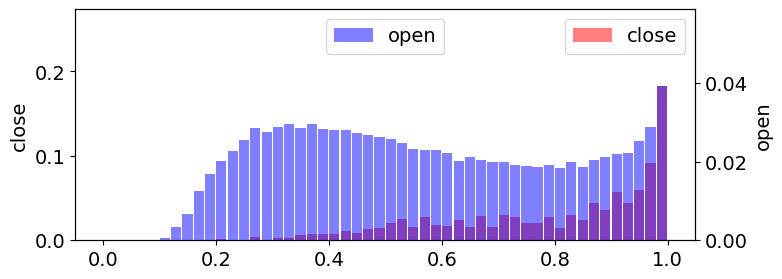}}
  \centerline{\footnotesize (o) ImageNet-LT, Group 0, $N_O=0$. }
\end{minipage}
\hfill
\begin{minipage}[b]{4cm}
  \centering
  \centerline{\includegraphics[width=4cm]{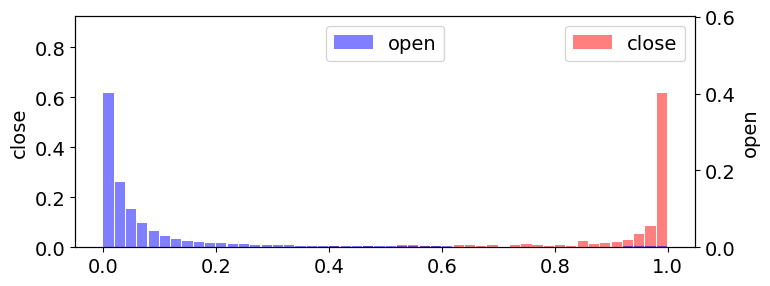}}
  \centerline{\footnotesize (p) ImageNet-LT, Group 0, $N_O=40$. }
\end{minipage}
\caption{The comparisons of distributions of the closed-set maximum probabilities predicted from both the closed-set and open-set data between introducing and not introducing open words into the proposed M-Tuning in the unknown detection experiments.}
\label{fig:max_close_prob_supp}
\end{figure*}

\textbf{Detailed Open-Set Recognition.} We evaluate the performance of our method for the detailed open-set recognition, in which both the known classes and unknown one need to be classified. In consistent with the literatures~\cite{neal2018open, yoshihashi2019classification, oza2019c2ae}, we set CIFAR10 as known, and ImageNet-crop, ImageNet-resize~\cite{russakovsky2015imagenet}, LSUN-crop, LSUN-resize~\cite{yu2015lsun} as 4 sets of unknown data. By setting the threshold $\tau_{max} \in [0.90, 0.92, 0.94, 0.96, 0.98]$ in Eq.~\ref{eq:thresh_pred}, the comprehensive performance of recognition on 11 classes (10 known classes in CIFAR10 and 1 unknown class) measured by mF1-score are compared in the Fig.~\ref{fig:mF1_score}.

Under the 16-shot setting, our method has an absolute advantage and is far greater than the prompt methods, i.e., CLIP+ZSL, CoOp and CoCoOp, in performance. Under the all-data setting, our method also exhibits superior performance than the OSR methods, i.e., ARPL and CE+, both in 4 sets of unknown data. Focusing on the dark line on the top which shows our method achieves stable performance under the all-data setting, it reveals that the known classes are almost confidently predicted with the maximum probability higher than 0.98, while the unknown classes are predicted with the closed-set maximum probability lower than 0.90 at least by our method. The result comparisons clearly prove that our method is effective in the detailed open-set recognition task by mitigating the label bias of prompt learning.

In conclusion, \emph{M-Tuning exhibits both parameter-efficiency and data-efficiency} by the above comparisons with newly constructed baselines. For one thing, it achieves the excellent performance on small datasets. For another thing, the proposed CTT successfully applies M-Tuning from small to large datasets for the best performance. Moreover, the outstanding results show that our method is effective in mitigating the label bias of prompt learning and also enables the prompts to be equipped with strong OSR ability.
\subsection{Effectiveness Validation of M-Tuning}
To validate the effectiveness of the proposed M-Tuning, we compare the results under the settings corresponding to Table~\ref{tab:unknown_detection_smalldata} and Table~\ref{tab:unknown_detection_largedata}, denoted as `OPTM', with no using open words. As the open score $S_{open}$ is definitely 0 as defined in Eq.~\ref{eq:open_score} without using open words, we compare the AUROC using MSP~\cite{hendrycks2016baseline} as the indicator in the Table~\ref{tab:no_OW}. 

When M-Tuning is performed without open words by setting $N_O=0$, the AUROC results are all lower than those under the optimal settings. The quantitative comparisons validate the effectiveness of M-Tuning in mitigating the label bias of prompt learning by introducing open words to simulate the open-set scenarios. 
\begin{table*}[t!]
    \scriptsize
	\centering
	\caption{AUROC comparisons  between using and not using open words in M-Tuning. AUROC is measured by adopting MSP~\cite{hendrycks2016baseline} as the indicator. ``OPTM'' refers to the optimal number settings of open words $N_O$ corresponding to Table~\ref{tab:unknown_detection_smalldata} and Table~\ref{tab:unknown_detection_largedata}.}
  \label{tab:no_OW}
  \begin{tabular}{c@{} |c@{} c@{} c@{} c@{} c@{} c@{} c@{} c@{} c@{}}
    \toprule
    \midrule
    Settings & \,\,\,CIFAR10\,\,\, & \,\,\,CIFAR+10\,\,\, & \,\,\,CIFAR+50\,\,\, & \,\,\,TinyImageNet\,\,\, & \,\,\,CIFAR100\,\,\, & \,\,\,ImageNet-100\,\,\, & \,\,\,ImageNet-200\,\,\, & \,\,\,ImageNet-LT\,\,\, \\
    \midrule
    \,\,\,$N_O = 0$ \,\,\,\,\,\,& 91.02 & 93.72 & 93.57 & 85.18 & 86.02 & 84.29 & 81.99 & 71.54 \\
    OPTM      & \textbf{94.63} & \textbf{96.11} & \textbf{95.47} & \textbf{86.79} & \textbf{88.38} & \textbf{97.78} & \textbf{94.39} & \textbf{77.34}\\
    \bottomrule
  \end{tabular}
\end{table*}
\begin{figure}[tb!]
\begin{minipage}[b]{4cm}
  \centering
  \centerline{\includegraphics[width=4.5cm]{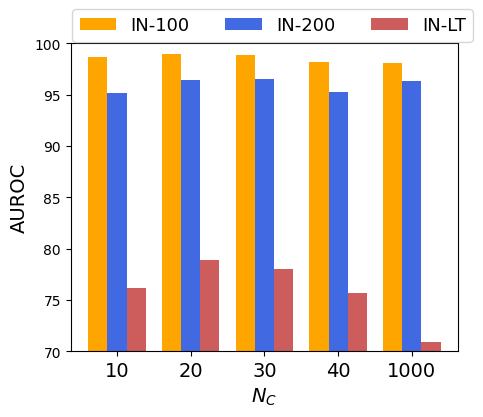}}
  \centerline{\footnotesize (a) AUROC.}
\end{minipage}
\hfill
\begin{minipage}[b]{4cm}
  \centering
  \centerline{\includegraphics[width=4.5cm]{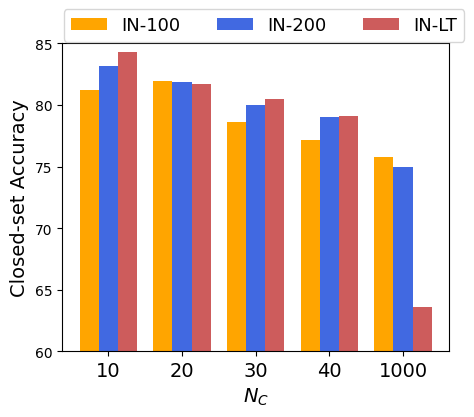}}
  \centerline{\footnotesize (b) Closed-set accuracy.}
\end{minipage}
\caption{Performance comparisons of the proposed M-Tuning with different number of classes $N_C$ in each group on large-scale datasets. `IN' denotes ImageNet.} 
\label{fig:CTT_verify}
\end{figure}
\begin{table*}[t!]
    \scriptsize
	\centering
	\caption{The AUROC comparisons under different prompt formats. ``pos" means the position of $[class]$ relative to the prompts.}
  \label{tab:prompt_format_auroc}
  \begin{tabular}{c@{} |c@{} c@{} c@{} c@{} c@{} c@{} c@{} c@{} c@{}}
    \toprule
    \midrule
    Settings & \,\,\,CIFAR10\,\,\, & \,\,\,CIFAR+10\,\,\, & \,\,\,CIFAR+50\,\,\, & \,\,\,TinyImageNet\,\,\, & \,\,\,CIFAR100\,\,\, & \,\,\,ImageNet-100\,\,\, & \,\,\,ImageNet-200\,\,\, & \,\,\,ImageNet-LT\,\,\, \\
    \midrule
    \,\,\,$L=10$, pos = front \,\, & 95.96 & 95.81 & 96.08 & \textbf{88.94} & 88.48 & 98.80 & 95.26 & 76.84 \\
    \,\,\,$L=10$, pos = \,end\,\,\, & 94.26 & 96.07 & 95.55 & 86.98 & 88.57 & 98.67 & \textbf{96.58} & 77.36 \\
    \,\,\,$L=15$, pos = \,mid\,\,\, & 94.38 & 96.40 & 95.46 & 87.14 & \textbf{89.56} & 98.84 & 95.79 & 74.12 \\
    \,\,\,$L=\,\,\,5$, pos = \,mid\,\,\, & \textbf{96.35} & \textbf{96.55} & 95.13 & 88.31 & 88.64 & \textbf{98.98} & 95.27 & 77.37 \\
    \,\,\,$L=10$, pos = \,mid\,\,\, & 96.29 & 96.28 & \textbf{96.17} & 87.30 & 89.51 & 98.93 & 96.47 & \textbf{78.89} \\
    \bottomrule
  \end{tabular}
\end{table*}
\begin{table*}[t!]
    \scriptsize
	\centering
	\caption{The closed-set accuracy comparisons under different prompt formats. ``pos" means the position of $[class]$ relative to the prompts.}
  \label{tab:prompt_format_acc}
  \begin{tabular}{c@{} |c@{} c@{} c@{} c@{} c@{} c@{} c@{} c@{} c@{}}
    \toprule
    \midrule
    Settings & \,\,\,CIFAR10\,\,\, & \,\,\,CIFAR+10\,\,\, & \,\,\,CIFAR+50\,\,\, & \,\,\,TinyImageNet\,\,\, & \,\,\,CIFAR100\,\,\, & \,\,\,ImageNet-100\,\,\, & \,\,\,ImageNet-200\,\,\, & \,\,\,ImageNet-LT\,\,\, \\
    \midrule
    \,\,\,$L=10$, pos = front \,\, & 95.40 & 95.55 & 95.98 & 89.20 & 87.20 & 78.14 & 80.19 & 81.94 \\
    \,\,\,$L=10$, pos = \,end\,\,\, & 95.11 & 95.83 & \textbf{96.38} & 90.20 & 87.25 & 80.20 & 80.80 & 82.24 \\
    \,\,\,$L=15$, pos = \,mid\,\,\, & 96.02 & 96.13 & 96.15 & 90.00 & \textbf{89.20} & 79.14 & 80.69 & \textbf{82.55} \\
    \,\,\,$L=\,\,\,5$, pos = \,mid\,\,\, & 96.22 & 95.93 & 96.35 & 89.60 & 85.55 & 77.30 & 79.42 & 81.42 \\
    \,\,\,$L=10$, pos = \,mid\,\,\, & \textbf{96.30} & \textbf{96.20} & 96.23 & \textbf{90.70} & 88.45 & \textbf{81.96} & \textbf{81.90} & 81.74 \\
    \bottomrule
  \end{tabular}
\end{table*}

In addition, for intuitively understanding the effectiveness of M-Tuning compared with standard prompts, that is no open words is introduced in the tuning phase, the distributions of closed-set maximum probabilities on each dataset without introducing open words (standard prompts, $N_O=0$) and introducing the optimal number of open words into M-Tuning are visualized in the Fig.~\ref{fig:max_close_prob_supp}. Without using open words, the closed-set maximum probability distributions of closed-set and open-set data exhibit significant overlap at the higher side, caused by the label bias of standard prompts. It shows that standard prompts could lead all testing samples to be predicted as closed-set classes with high probabilities. After introducing the open words, the distributions of open-set and closed-set data are clearly separated by the proposed M-Tuning. Taking CIFAR+50 as an example, when tuning without using open words, $15\%$ open-set samples are predicted with the closed-set maximum probabilities distributing at the highest side. With the proposed M-Tuning, the portion drops from $15\%$ to lower than $2\%$ significantly. Closed-set data is predicted accurately, while open-set data is predicted with much lower probabilities on the closed-set classes. The qualitative visualization results validate the effectiveness of the proposed M-Tuning in mitigating the label bias of prompt learning, contributing to the outstanding OSR performance.

\subsection{Effectiveness Validation and Efficiency Analysis} of CTT
To show the effectiveness of the proposed CTT on large-scale datasets, we compare the results by setting the number of classes in each group $N_C \in [10, 20, 30, 40, 1000]$ in the Fig.~\ref{fig:CTT_verify}. Because the numbers of closed-set classes of ImageNet-100, ImageNet-200 and ImageNet-LT are 100, 200, 1000, respectively, setting $N_C=1000$ refers to that M-Tuning and inference are performed directly on large datasets without grouping according to Eq.~\ref{eq:group_rule}, that is $G=1$. The closed-set accuracy under the settings $N_C \in [10, 20, 30, 40]$ are almost better than those under $N_C=1000$ ($G=1$). In particular, the closed-set accuracy results are the lowest when $N_C=1000$ ($G=1$). The comparisons verify the effectiveness of the proposed CTT, which decomposes M-Tuning directly on large datasets into multiple independent M-Tuning on groups with fewer classes, contributing to the improved performance on large datasets. Besides, from the ablations on the number of classes in each group $N_C$, the highest AUROC results on the three datasets are achieved by setting $N_C=20$.

On the contrary, more groups would result in more prompts and tuning costs. In the proposed M-Tuning, the computation complexity for each single group is fixed. After leading the CTT strategy, the computation complexity is $G$ times that of a single group. According to the results in the Fig.~\ref{fig:CTT_verify}, CTT mainly affects the accuracy of closed-set classification, compared with the AUROC. To make a balance between the computation efficiency and the performance, the trade-off setting is to set $N_C$ as 40. 

\subsection{Ablation on Prompt Formats}
\label{sec:ablation_prompt_formats}
In the above main experiments, the prompt length is set as 10 and the $[class]$ is put in the middle place of the prompt. In this part, we deliver the ablation study on different formats of prompts. We set the prompt length $L$ within $[5, 10, 15]$, and the positions that the $[class]$ relative to the prompts as front, middle (mid), end for comprehensive comparison. The results measured by AUROC and closed-set accuracy under different prompt formats are given in the Table~\ref{tab:prompt_format_auroc} and Table~\ref{tab:prompt_format_acc}.

The comparisons show that the comprehensive performance advantage on the AUROC and closed-set accuracy is achieved by setting $L=10$ and putting the $[class]$ in the middle of prompts. It reveals that too long or too short prompts are inferior, and putting the $[class]$ in the middle place could help in learning effective prompts.

\begin{table}[t!]
    \scriptsize
	\centering
	\caption{The ablation study on the prompt initialization under all-data setting. ``Gaussian" and ``Text" represent the prompts are initialized randomly by Gaussian distribution and the text respectively.}
  \label{tab:ini_ablation}
  \begin{tabular}{c@{} |c@{} c@{} | c@{} c@{} }
    \toprule
    \midrule
    \multirow{2}{*}{\,\,Datasets\,\,} &  \multicolumn{2}{c|}{AUROC} &  \multicolumn{2}{c}{Closed-set Acc} \\
    \cline{2-5}
    
               & \,\,\,\,Gaussian\,\,\,\, & \,\,\,\,\,\,\,\,Text\,\,\,\,\,\,\,\,  & \,\,\,\,Gaussian\,\,\,\, & \,\,\,\,\,\,\,\,Text\,\,\,\,\,\,\,\,  \\
    \midrule
    CIFAR10    & \,\,\,\,\textbf{96.29}\,\,\,\, & \,\,\,\,94.54\,\,\,\,  & \,\,\,\,\textbf{96.30}\,\,\,\,  & \,\,\,\,96.00\,\,\,\, \\
    CIFAR+10   & \textbf{96.28} & 94.88  & \textbf{96.20}  & 95.66 \\
    CIFAR+50   & 96.17  & \textbf{96.20} & 96.23  & \textbf{96.25}  \\
    \,\,\,\,TinyImageNet\,\,\,\,     & \textbf{87.30}   & 85.52  & \textbf{90.70}  & 90.50 \\
    CIFAR100    & \textbf{89.51} & 88.21   & \textbf{88.45}  & 86.25 \\
    \bottomrule
  \end{tabular}
\end{table}

\subsection{Ablation on Prompt Initialization}
To show the influence caused by the initialization, inspired by CoOp~\cite{zhou2022learning} and CoCoOp~\cite{zhou2022conditional}, we conduct the ablation study by initializing prompts using the text ``this is a photo of \_\_ in the open set scenario.". The results are given in the Table~\ref{tab:ini_ablation}. The Gaussian initialization performs slightly better. From this, \textit{different initialization could lead to different results}, thus we propose to use the Gaussian initialization for convenience. 
\begin{table}[t!]
    \scriptsize
	\centering
	\caption{AUROC comparisons of the proposed M-Tuning with different number of open words $N_O$.}
  \label{tab:OW_ablation}
  \begin{tabular}{c@{} |c@{} c@{} c@{} c@{} }
    \toprule
    \midrule
    Datasets & \,\,\,10\,\,\, & \,\,\,20\,\,\, & \,\,\,40\,\,\, & \,\,\,60\,\,\,\\
    \midrule
    CIFAR10        & \,\,\,\,\,\,95.74\,\,\,\,\,\, & \,\,\,\,\,\,\textbf{96.29}\,\,\,\,\,\, & \,\,\,\,\,\,96.10\,\,\,\,\,\, & \,\,\,\,\,\,95.84\,\,\,\,\,\, \\
    CIFAR+10       & 95.64 & \textbf{96.28} & 95.83 & 96.16 \\
    CIFAR+50       & \textbf{96.17} & 95.82 & 95.58 & 95.97 \\
    \,\,\,TinyImageNet\,\,\,\,\,\,  & 86.08 & \textbf{87.30} & 86.51 & 86.43 \\
    CIFAR100       & 87.68 & \textbf{89.51} & 87.08 & 87.30 \\
    ImageNet-100   & 98.28 & 98.54 & 98.55 & \textbf{98.93} \\
    ImageNet-200   & 94.45 & 95.14 & 96.01 & \textbf{96.47} \\
    ImageNet-LT    & 74.44 & 75.60 & \textbf{78.89} & 77.06 \\
    \bottomrule
  \end{tabular}
\end{table}
\begin{table*}[t!]
    \scriptsize
	\centering
	\caption{Performance comparisons with different grouping strategies of CTT on large-scale datasets.}
  \label{tab:grouping}
  \begin{tabular}{c@{} |c@{} c@{} c@{}|c@{} c@{} c@{}}
    \toprule
    \midrule
    \multirow{2}{*}{\,\,Strategies\,\,\,\,\,\,} &  \multicolumn{3}{c|}{AUROC} &  \multicolumn{3}{c}{Closed-set Accuracy} \\
    \cline{2-7}
                            & \,\,\,ImageNet-100\,\,\, & \,\,\,ImageNet-200\,\,\, & \,\,\,ImageNet-LT\,\,\, & \,\,\,ImageNet-100\,\,\, & \,\,\,ImageNet-200\,\,\, & \,\,\,ImageNet-LT\,\,\, \\
    \midrule
    Random & 98.83 & 96.04 & 78.73 & 78.12 & \textbf{84.65} & \textbf{84.04} \\
    Semantics    & 98.90 & 95.13 & 74.82 & 72.16 & 69.45 & 64.34 \\
    \,\,\,WNID Order\,\,\,   & \textbf{98.93} & \textbf{96.47} & \textbf{78.89} & \textbf{81.96} & 81.90 & 81.74 \\
    \bottomrule
  \end{tabular}
\end{table*}
\begin{table*}[t!]
    \scriptsize
	\centering
	\caption{Performance comparisons between using and not using the true names of open-set data in M-Tuning.}
  \label{tab:true_open}
  \begin{tabular}{c@{} |c@{} |c@{} c@{} c@{} c@{} c@{} c@{} c@{} c@{} c@{}}
    \toprule
    \midrule
    Metrics & \,Settings\,\,\,\, & \,\,\,CIFAR10\,\,\, & \,\,\,CIFAR+10\,\,\, & \,\,\,CIFAR+50\,\,\, & \,\,\,TinyImageNet\,\,\, & \,\,\,CIFAR100\,\,\, & \,\,\,ImageNet-100\,\,\, & \,\,\,ImageNet-200\,\,\, & \,\,\,ImageNet-LT\,\,\, \\
    \midrule
    \multirow{2}{*}{\,\,AUROC \,\,\,}        & \,True Open\,\,\,\,       & 95.36 & 96.00 & 96.13 & 86.99 & 88.69 & 93.48 & 90.52 & \textbf{81.68} \\
                                             & \,No Using\,\,\,\,   & \textbf{96.29} & \textbf{96.28} & \textbf{96.17} & \textbf{87.30} & \textbf{89.51} & \textbf{98.93} & \textbf{96.47} & 78.89 \\
    \midrule
    \multirow{2}{*}{\,\,Closed-set Acc\,\,\,}& \,True Open\,\,\,\,       & \textbf{96.32} & 96.00 & 95.48 & 89.30 & \textbf{88.55} & 80.18 & 81.64 & 74.23 \\
                                             & \,No Using\,\,\,\,   & 96.30 & \textbf{96.20} & \textbf{96.23} & \textbf{90.70} & 88.45 & \textbf{81.96} & \textbf{81.90} & \textbf{81.74} \\
    \bottomrule
  \end{tabular}
\end{table*}

\subsection{Ablation on the Number of Open Words $N_O$ in M-Tuning}
\label{sec:OW_ablation}
To show how the number of open words in M-Tuning affects the performance, we give the AUROC result comparisons by setting $N_O \in [10, 20, 40, 60]$ in the Table~\ref{tab:OW_ablation}. The AUROC results change within a small range with different $N_O$. The largest AUROC gap is only $3.29$ on ImageNet-LT dataset when setting the number of open words $N_O$ as 20 and 40. The results reveal that our method could mitigate the label bias with less sensitivity to the number of open words. The less sensitivity to the number of open words is mainly brought by the expanded label space for prediction. Considering that the number of known classes in the small datasets ranges from 4 to 20, and the large datasets is divided into multiple groups with each group consisting of 20 known classes, setting $N_O \in [10, 20, 40, 60]$ for each prompt could expand the label space to at least two times larger, thus effectively mitigate the label bias.

\subsection{Ablation on Grouping Strategies of CTT}
\label{sec:grouping_ablation}
In this section, we perform the ablation study on the grouping strategies of CTT on large datasets, including grouping by the WNID order in above main experiments, grouping by random selection, and grouping by semantic similarities.

In the WNID grouping strategy, the class names are sorted by their WordNet IDs. In the random grouping strategy, the class names are sorted randomly. In the semantic grouping strategy, we sort the classes by the similarities of their text embeddings extracted by the text encoder of CLIP~\cite{radford2021learning}. We select the first class in closed-set labels as the start in the semantic order, the class with the highest similarity to the previous one is successively appended to the semantic order list. As a result, any two adjacent categories in the list are the ones with the highest semantic similarity. After the sorting process, the groups are constructed by selecting the continuous segment of the class name list one by one.

The results under different grouping strategies are compared in the Table~\ref{tab:grouping}. It shows that grouping by the WNID order achieves the best AUROC results and competitive closed-set accuracy. To make a deep investigation on this phenomenon, taking ImageNet100 as an example, we divide the closed-set with 100 classes into 5 groups with 20 classes per group using the three strategies. We feed the class names with learned prompts into the text encoder of CLIP. For each group, we calculate the averaging similarities among the text embeddings of the 20 classes. Then, we calculate the averaging value on the 5 groups. The averaging similarities caused by WNID, random and semantic grouping strategies are 64.07, 64.57 and 68.92, respectively. The WNID and semantic strategies lead to the lowest and highest similarities. This demonstrates that the WNID order contributes to the optimal inter-class split both within and across all groups, by which the classes within a group are easily classified, the optimal prompts are easily selected with less confusion. However, the closed-set accuracy results under the semantics grouping strategy are less satisfactory. The classes in each group are quite similar to each other, tending to bring the prediction confusion. 

\subsection{Necessity of Knowing True Names of Open-set Data}
\label{sec:necessity}
The OOD method~\cite{fort2021exploring} leverages the true names of outlier classes for outlier exposure. However, in the realistic open-set scenarios, we have no knowledge of them. To validate whether we need to know the true names of open-set data, we perform the comparisons between knowing and unknowing the true names of the open-set data. The settings using and not using true names of open-set data are depicted as ``True Open" and ``No Using", respectively. We denote the optimal setting of the number of open words in the Table~\ref{tab:OW_ablation} as $N_O^{opt}$ uniformly. Specifically, there are two cases in M-Tuning when knowing the true names in the ``True Open" setting: i) if the number of open-set classes is fewer than $N_O^{opt}$, e.g., there are 4 open-set classes in the setting of CIFAR10 while $N_O^{opt}$ is 20, all the open-set classes are integrated into M-Tuning, in addition with other open words to achieve $N_O^{opt}$ words in total; ii) if the number of open-set classes is no fewer than $N_O^{opt}$, we directly select $N_O^{opt}$ names from all the open-set classes. 
\begin{table*}[t!]
    \scriptsize
	\centering
	\caption{The average similarities between the text embeddings extracted by the text encoder. The text encoder takes the learned prompts filled with the open words and closed-set classes as input.}
  \label{tab:word_relationship}
  \begin{tabular}{c@{} |c@{} c@{} c@{} c@{} c@{} c@{} c@{} c@{} c@{}}
    \toprule
    \midrule
    Settings & \,\,\,CIFAR10\,\,\, & \,\,\,CIFAR+10\,\,\, & \,\,\,CIFAR+50\,\,\, & \,\,\,TinyImageNet\,\,\, & \,\,\,CIFAR100\,\,\, & \,\,\,ImageNet-100\,\,\, & \,\,\,ImageNet-200\,\,\, & \,\,\,ImageNet-LT\,\,\, \\
    \midrule
    \,\,\,True Open \,\, & 15.68 & 23.01 & \textbf{17.99} & \textbf{19.95} & 17.73 & \textbf{62.45} & \textbf{50.28} & \textbf{53.24} \\
    \,\,\,No Using \,\, & \textbf{18.35} & \textbf{23.23} & 9.85 & 18.01 & \textbf{18.34} & 58.92 & 41.13 & 50.93 \\
    \bottomrule
  \end{tabular}
\end{table*}

From Table~\ref{tab:true_open}, the results of M-Tuning without using the true names of the open-set data are almost better than those with using the true names. It proves that it is not necessary to know the true names of open-set data for M-Tuning, which is also more consistent with the challenging image recognition tasks in reality, i.e., we do not know any information about the unseen data, even their names.

\section{Investigation of Open Words}
\label{sec:Further_Investigation}
To further investigate the effect caused by the open words, we perform two sets of experiments for a general principle of selecting open words. The first setting is to select open words manually by introducing the true names of open-set data as the open words, the results have been delivered in the Sec.~\ref{sec:necessity}. The second setting is to select open words according to their semantic similarities with the closed-set classes.

\subsection{Manual Selection}
In this setting, open words are selected manually as the true names of the open-set data. According to the results in the Table~\ref{tab:true_open}, the comprehensive performance achieved under the ``True Open" setting is slightly inferior to that under the ``No Using" setting. In addition, we compare the average text similarities between the selected open words and closed-set classes based on the validation of the necessity of knowing the true names of the open-set data in Sec.~\ref{sec:necessity}. By filling the closed-set classes and open words into the prompts, the similarities between them are calculated on the embeddings extracted by the text encoder of CLIP, as shown in the Table~\ref{tab:word_relationship}. 

Combining the comparisons of the performance in the Table~\ref{tab:true_open} and the word similarities in the Table~\ref{tab:word_relationship}, we could observe that \emph{the lower similarities between the open words and close-set classes may contribute to better performance}, especially the better closed-set classification accuracy. 

\subsection{Semantic Similarity based Selection}
In this setting, open words are selected according to the semantic similarities. For each dataset, the number of selected open words is the same as the optimal setting of the number of open words in the Table~\ref{tab:OW_ablation}. We use the small datasets for efficient validation. In detail, by filling the closed-set classes and a large number of words from the WordNet into the prompt, i.e., ``\textit{a photo of a \{\}.}", the semantic similarities among the closed-set classes and other words are calculated. After sorting the similarities, a set of open words with high or low similarities to the closed-set classes are obtained. The selected open words are elaborated in the Table~\ref{tab:semantic_open_words}.

By leveraging the selected open words with high or low semantic similarities to the closed-set classes for M-Tuning, the results are delivered in the Table~\ref{tab:high_low_semantic_OW}. Better performance could be almost achieved by selecting the open words with lower semantic similarities to the closed-set classes. This observation is consistent with the one from the experiments of manual selection.
\begin{table*}[t!]
    \scriptsize
	\centering
	\caption{The selected open words that exhibit high and low semantic similarities to the closed-set classes.}
  \label{tab:semantic_open_words}
  \begin{tabular}{ c@{} | c@{} |c@{}}
    \toprule
    \midrule
    Datasets & \,\,\,\,Semantic Similarity\,\,\,\, & Selected Open Words \\
    \midrule
    \rule{0pt}{8pt} \multirow{2}{*}{\,\,\,\,CIFAR10\,\,\,\,} & \,\,\,\,High\,\,\,\, & \makecell{cow\_pony, cross\_bun, chick, great\_ape, star, ape, pigeon, tack, calf, pet, mule, cock,\\ container, electric\_hammer, wallet, doctor, inhabitant, car\_train, radicchio, representative.} \\
    \cline{2-3} \rule{0pt}{12pt} & Low & \makecell{cowboy\_hat, brake, even\_toed\_ungulate, artichoke, token, crab\_apple, cockfighting, belted\_kingfisher, German\_short\_haired\_pointer, \\ jersey, floor, berlin, pilot\_whale, washboard, charger, peach\_ice\_cream, cream\_soda, butterfly\_orchid, family\_room, chervil.}\\
    \midrule
    \rule{0pt}{8pt} \multirow{2}{*}{\,\,\,\,CIFAR+10\,\,\,\,} & High & \makecell{mascot, rocket, building, entree, cab, Seven\_Wonders\_of\_the\_Ancient\_World, fox, tom, toy, bitter, mammal,\\ chick, automobile\_horn, near\_beer, separate, stock, trailer\_truck, Xerox, automatic\_firearm, middle\_aged\_man.} \\
    \cline{2-3} \rule{0pt}{12pt} & Low & \makecell{swan, garambulla, firebrick, lingerie, Bouvier\_des\_Flandres, breadfruit, coneflower, red\_drum, deodar, brown\_rat,\\ automat, star, bodyguard, astilbe, great\_St\_Johns\_wort, ladder\_back, arete, defile, Swan\_River\_daisy, white\_fir.}\\
    \midrule
    \rule{0pt}{8pt} \multirow{2}{*}{\,\,\,\,CIFAR+50\,\,\,\,} & High & \makecell{primate, device, turkey, rodent, ape,\\ toy\_dog, mate, White, foe, gopher.} \\
    \cline{2-3} \rule{0pt}{12pt} & Low & \makecell{waterside, Bordeaux, orchard\_oriole, beef\_stew,\\ grenadine, powder, porch, cannon, binder, sieve.}\\
    \midrule
    \rule{0pt}{8pt} \multirow{2}{*}{\,\,\,\,TinyImageNet\,\,\,\,} & High & \makecell{weapon, keyboard, pig\_bed, bridge, earthenware, convenience\_store, folding\_chair, tie\_rack, diocesan, suspension\_bridge,\\ lunch, sea\_slug, hors\_doeuvre, undergarment, diet, amphibian, wing\_chair, hedgehog, Atlantic\_salmon, vegetable.} \\
    \cline{2-3} \rule{0pt}{12pt} & Low & \makecell{lectern, hotel\_casino, double\_cream, clavier, plant, dress\_hat, junction, jasmine, kingfish, garment\_bag,\\ Polynesian, eraser, shuffleboard, stolon, sloth\_bear, safety\_belt, king, snack\_food, pistil, soccer\_player.}\\
    \midrule
    \rule{0pt}{8pt} \multirow{2}{*}{\,\,\,\,CIFAR100\,\,\,\,} & High & \makecell{contact, horn, bread\_dough, tumbleweed, ox, space\_bar, pill\_bottle, understudy, stagecoach, colt, rifle,\\ common\_mosquito, boar, radiotelephone, aoudad, motorcyclist, dresser, croaker, clown\_anemone\_fish, body.} \\
    \cline{2-3} \rule{0pt}{12pt} & Low & \makecell{sumo, hard\_hat, apple, woofer, Charolais, muscovy\_duck, tawny\_eagle, palo\_verde, satin\_bowerbird, foxhole,\\ Herero, globe\_lily, sour\_bread, felt, thumb, hornbeam, false\_scorpion, hand\_tool, Saint\_Emilion, carving\_fork.}\\
    \bottomrule
  \end{tabular}
\end{table*}
\begin{table*}[t!]
    \scriptsize
	\centering
	\caption{Performance comparisons between using the open words with high and low semantic similarities to the closed-set categories for M-Tuning.}
  \label{tab:high_low_semantic_OW}
  \begin{tabular}{c@{} |c@{} |c@{} c@{} c@{} c@{} c@{} c@{}}
    \toprule
    \midrule
    Metrics & \,Sematic Similarity\,\,\,\, & \,\,\,CIFAR10\,\,\, & \,\,\,CIFAR+10\,\,\, & \,\,\,CIFAR+50\,\,\, & \,\,\,TinyImageNet\,\,\, & \,\,\,CIFAR100\,\,\,  \\
    \midrule
    \multirow{2}{*}{\,\,AUROC \,\,\,}        & \,High\,\,\,\,       & 95.04 & 97.35 & 96.12 & \textbf{87.72} & \textbf{88.21}  \\
                                             & \,Low\,\,\,\,        & \textbf{96.52} & \textbf{97.61} & \textbf{96.83} & 87.69 & 87.80  \\
    \midrule
    \multirow{2}{*}{\,\,Closed-set Acc\,\,\,}& \,High\,\,\,\,       & 96.20 & 96.25 & 96.38 & 90.30 & 89.65  \\
                                             & \,Low\,\,\,\,        & \textbf{96.78} & \textbf{96.45} & \textbf{96.63} & \textbf{90.70} & \textbf{90.85}  \\
    \bottomrule
  \end{tabular}
\end{table*}

\subsection{Analysis of Open Words Selection}
According to the above experimental results, open words with lower similarities to the closed-set classes could result in regularized predictions, and the probabilities on the open words are clearly different from those on the closed-set classes. By this, the closed-set images are predicted with extremely high probabilities, while the probabilities on the open words are quite low. On the contrary, open words with high similarities to the closed-set classes would result in confusion predictions, as the probabilities on the open words and closed-set classes are similar. As a result, the prompts could still select words similar to the closed-set classes, even exactly the closed-set classes, as the final predictions, causing the unsatisfaction in mitigating the label bias. \emph{The observation further proves that in practical applications of the proposed M-Tuning, we do not need to know the true names of open-set data.}

Moreover, based on the above experiments, \textit{our method is thus proved to be not only applicable with the assistance of the WordNet}. To simulate the open-set scenario, a database containing lots of words, e.g., the WordNet, is expected for us to select words and expand the label set in M-Tuning. It may also work when we use other databases, because \textit{the performance is validated to be related to the similarities between the open words and the closed-set classes}. A general principle is concluded that selecting open words with lower semantic similarity to the closed-set classes could lead the better performance.

\section{Conclusions}
In this paper, facing the realistic recognition task that we do not know any information even names of the unseen classes, we propose the M-Tuning. It mitigates the label bias, that current prompt methods exhibits, by introducing open words into tuning to simulate the open-set scenarios. To the best of our knowledge, this is the first vision-language prompt learning method targeting at the challenging open-set scenario. In addition, to achieve better performance on large datasets as that on small datasets with fewer classes, the Combinatorial Tuning and Testing (CTT) strategy is devised. Moreover, seeing the lack of VL-based OSR baselines, especially for prompt methods, we construct new baselines for fair comparisons. Our method achieves the best performance both on small and large datasets compared with baseline methods. Extensive ablation studies validate the effectiveness of each component of our method.

\section*{Acknowledgement}
This work was partly supported by National Science Foundation of China under Grant NSFC 62222607 and NSFC 62476188, Shanghai Municipal Science and Technology Major Project (2021SHZDZX0102).
\bibliographystyle{IEEEtran}
\bibliography{egbib.bib}

\end{document}